\pdfoutput=1

\documentclass[11pt]{article}

\usepackage{EMNLP2023}

\usepackage{times}
\usepackage{latexsym}

\usepackage[T1]{fontenc}

\usepackage[utf8]{inputenc}
\usepackage{inconsolata}

\usepackage{microtype}

\usepackage{enumitem}
\usepackage{comment}
\usepackage{url}
\usepackage{arydshln}
\usepackage{graphicx}
\usepackage{booktabs}
\usepackage{pifont}
\usepackage{graphicx}
\usepackage{subcaption}

\usepackage{algorithm}
\usepackage[noend]{algpseudocode}
\usepackage{xcolor, soul}
\usepackage{multirow, makecell}

\usepackage{listings}

\usepackage{adjustbox} 
\usepackage{amsmath} 

\definecolor{Background}{RGB}{217,245,203}
\sethlcolor{Background}

\definecolor{Back_Purple}{RGB}{238,215,236}
\definecolor{Back_Yellow}{RGB}{250,239,197}

\newif\ifcomments
\commentstrue 
\ifcomments
    \newcommand\tw[1]{\textcolor{magenta}{[TW: #1]}}
    \newcommand\amit[1]{\textcolor{blue}{[AS: #1]}}
    \newcommand\oren[1]{\textcolor{red}{[OG: #1]}}

\else
    \providecommand{\tw}[1]{}
    \providecommand{\amit}[1]{}
    \providecommand{\oren}[1]{}
    \providecommand{\yevgeni}[1]{}
\fi

\newcommand\bench{\textsc{WikiTabGen}}

\title{Generating Tables from the Parametric Knowledge of Language Models}

\author{\makecell{Yevgeni Berkovitch$^{1}$~~~~~ Oren Glickman$^{1}$~~~~~ Amit Somech$^{1}$~~~~~ Tomer Wolfson$^{2}$ } \\ 
$^{1}$Bar-Ilan University\hspace{5mm}
$^{2}$Tel Aviv University\hspace{5mm}\\ 
\texttt{\small\makecell{taoberkovitch@gmail.com}} 
\texttt{\small\makecell{\{oren.glickman, somecha\}@cs.biu.ac.il}} \texttt{\small\makecell{tomerwol@mail.tau.ac.il}}}

\begin{document}
\maketitle

\begin{abstract}
We explore generating factual and accurate tables from the parametric knowledge of large language models (LLMs). While LLMs have demonstrated impressive capabilities in recreating knowledge bases and generating free-form text, we focus on generating structured tabular data, which is crucial in domains like finance and healthcare.
We examine the table generation abilities of four state-of-the-art LLMs: GPT-3.5, GPT-4, Llama2-13B, and Llama2-70B, using three prompting methods for table generation: (a) full-table, (b) row-by-row; (c) cell-by-cell. For evaluation, we introduce a novel benchmark, \bench\ which contains 100 curated Wikipedia tables. Tables are further processed to ensure their factual correctness and manually annotated with short natural language descriptions. 
Our findings reveal that table generation remains a challenge, with GPT-4 reaching the highest accuracy at 19.6\%. Our detailed analysis sheds light on how various table properties, such as size, table popularity, and numerical content, influence generation performance. This work highlights the unique challenges in LLM-based table generation and provides a solid evaluation framework for future research. Our code, prompts and data are all publicly available.\footnote{\url{https://github.com/analysis-bots/WikiTabGen}}
\end{abstract}
\section{Introduction} 
\label{sec:introduction}

\begin{figure}[t]\setlength{\belowcaptionskip}{-8pt}
  \centering
  \includegraphics[clip, width=0.45\textwidth]{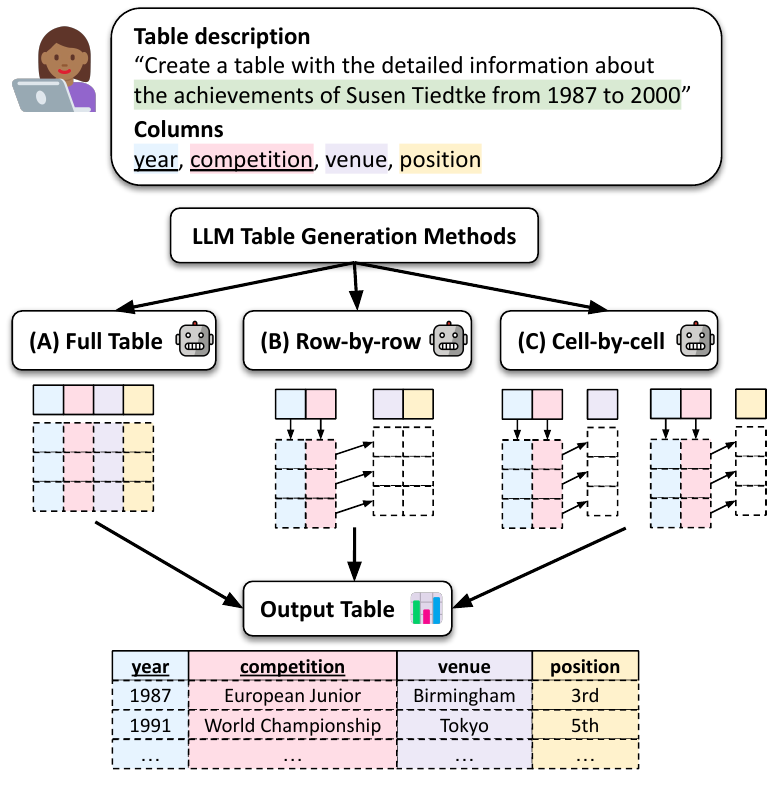}
  \caption{An example LLM-based table generation task, using three alternative prompting methods.}
  \label{fig:overview}
\end{figure}


Large language models (LLMs) \cite{NEURIPS2020_1457c0d6, Chowdhery2022PaLMSL,Kadavath2022LanguageM, Touvron2023LLaMAOA} are pre-trained on vast amounts of text which include scientific articles, encyclopedic knowledge, and importantly, tabular data \cite{Elazar2023WhatsIM,Fang2024LargeLM}.
Therefore, during pre-training, LLMs encode a significant amount of factual information in their parameters. Past work has shown that, based on their parametric knowledge, LLMs have the potential to elicit fairly accurate facts for tasks such as recreating knowledge bases (KBs) \cite{petroni2019language, AlKhamissi2022ARO, Cohen2023CrawlingTI} and generating Wikipedia-like articles \cite{Shao2024AssistingIW}.
Given the promising results on free-form text and knowledge bases, our work explores the task of \emph{generating tabular data} from the parametric knowledge of LLMs. Tabular data, being a structured representation, is widespread in real-world sources and, unlike free-form text, is typically used to represent large amounts of complex information. Tabular data is essential in sectors such as finance and healthcare \cite{Chen2021FinQAAD, Johnson2016MIMICIIIAF}, where data is further analyzed using statistical methods, and visualization tools \cite{Berant2018ExplainingQO, Shen2021TowardsNL}. 

A key research question, is to what extent can the parametric knowledge of LLMs be used to generate factual and accurate tables. To this end, our work focuses on evaluating the capability of LLMs to generate tabular data, given an input description of the table's content and a list of its columns. Past research has focused on tabular reasoning over existing input tables \cite{Pasupat2015CompositionalSP, Chen2020OpenQA,Chen2021FinQAAD}, or on eliciting factual knowledge from LLMs in the context of KBs or free-form text \cite{petroni2019language, Mallen2022WhenNT}. In contrast to these works, generating tables from LLMs' parametric knowledge presents unique challenges, as it requires understanding the table structure and perform long-form reasoning \cite{shaham-etal-2022-scrolls} on potentially hundreds of entries.

To address the problem of long-form table generation we implement three distinct prompting methods: (a) generating the \emph{full table} all at once, (b) \emph{row-by-row} generation, and (c) \emph{cell-by-cell} generation. The latter two methods use a modular prompting approach \cite{khot2023decomposed}, where one instance of the LLM is prompted to generate table \textit{keys}, i.e. unique identifiers of the table's entries, while another is prompted to generate either a full table row (row-by-row) or single table cell (cell-by-cell). 
Fig.~\ref{fig:overview} presents an example flow of LLM-based table generation in our setting. The target table is described as \textit{``the achievements of (long jumper) Susen Tidtke from 1987 to 2000''} and has four table columns:\texttt{year}, \texttt{competition}, \texttt{venue} and \texttt{position}. The first two columns (underlined) are the \emph{key} columns of the table. We then use either the full-table, row-by-row, or cell-by-cell method in order to generate the target table.

To evaluate the table generation capabilities of LLMs, we introduce the \bench benchmark. Each instance consists of a table description, a list of columns and the ground-truth table. 
We build a robust evaluation set by processing 100 Wikipedia tables, avoiding tables that are incomplete or contain subjective information. While \bench\ has 100 tables overall, we note that the average number of tokens per table is 1,457 tokens, significantly larger than in past tabular generation tasks \cite{Parikh2020ToTToAC, nan-etal-2022-fetaqa}.
\bench\ contains diverse tables with respect to several properties, namely: their size (number of rows, columns and cells); the portion of numerical content (numbers and dates); and the table popularity, indicating its prevalence in the LLM pre-training data \cite{Mallen2022WhenNT}. Our experiments and subsequent analysis describe how each of these properties affects table generation performance. 

For our table generation experiments we evaluated four popular LLMs: GPT-3.5, GPT-4, Llama2-13B, and Llama2-70B \cite{Touvron2023Llama2O, Achiam2023GPT4TR}. Our results reveal that generating tables from the parametric knowledge of LLMs remains an open problem, as our top-performing model GPT-4, with row-by-row prompting, achieves an F1 score of 19.6\% (\S\ref{sec:experiments}).

Our main contributions are as follows: (1) We formulate the problem of generating tables from the parametric knowledge of LLMs; (2) We implement and evaluate three prompt-based table generation methods; (3) We introduce \bench, a new table generation benchmark; and (4) We conduct an extensive analysis of the table generation performance of state-of-the-art LLMs. Our paper highlights key challenges in table generation, with factors such as table size, numerical values, and popularity all playing a significant role in LLM performance. 
We hope that our methods, analysis, and evaluation benchmark will encourage future research on tabular data generation with LLMs.







\begin{figure*}[t]\setlength{\belowcaptionskip}{-8pt}
  \centering
  \includegraphics[clip, width=0.99\textwidth]{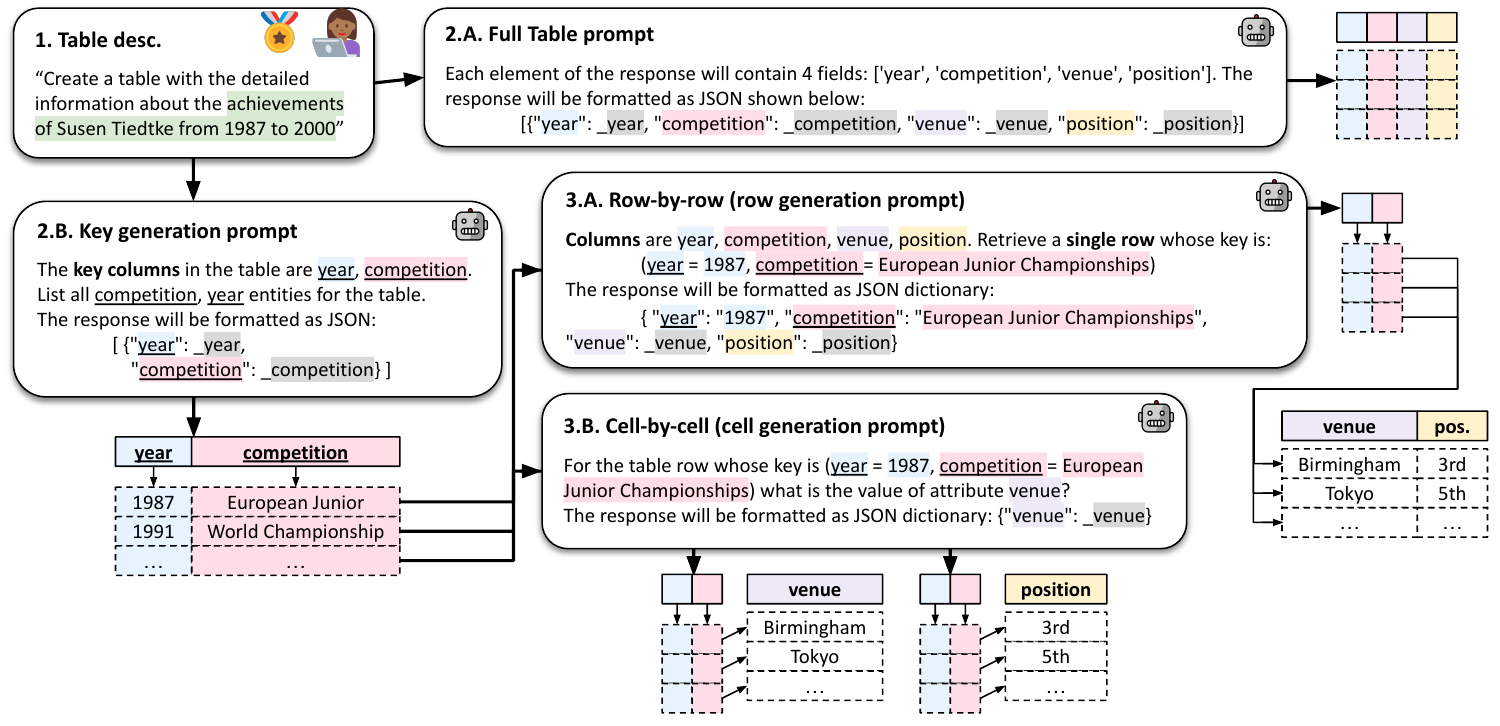}
  \caption{An overview of our three separate prompting methods for table generation, given a short user description and table metadata (Fig.~\ref{fig:overview}): (2.A) Full table directly generates the table given the user desc. and its columns; (2.B) Key-generation is used in both the row-by-row and cell-by-cell methods; (3.A) Row-by-row generates a table row given a unique key value, e.g. (1987, EU Junior Championship); (3.B) Cell-by-cell generates a single table cell given a key value and specific target column e.g. \textit{venue} $\rightarrow$ \textit{Birmingham}.} 
  \label{fig:architecture}
\end{figure*}
\section{Problem Definition}
\label{sec:problem}
Given a short user description, our goal is to generate a corresponding table with factually accurate knowledge. 
A relational table~\cite{codd1990relational} $T=(R,C)$ comprises a set of rows $R=\{r_1,r_2,\dots\}$ and a set of columns $C = \{c_1,c_2\dots\}$. A table cell, denoted $r[c]$, contains the value of column $c$ in row $r$. Key columns are a subset $C_k \subset C$ that uniquely define each entry (row) in $T$ and the corresponding cells do not contain null or empty values. For example, the table in Fig.~\ref{fig:ex_table} has the columns \texttt{year} and \texttt{competition} as its keys. Each table entry such as \texttt{venue}, corresponds to a unique \texttt{year}, \texttt{competition} pair.

Provided with a table description $d$ and a list of desired table columns $C$, our task is to generate a corresponding table $T(\hat{R},C)$, where the generated rows $\hat{R}$ contain factually accurate information.

An example problem is shown in Fig.~\ref{fig:overview}. The table description is \textit{``Achievements of Susen Tiedke from 1987 to 2000''}, while the target columns are: 
\textit{year}, \textit{competition},\textit{venue}, and \textit{position}. The first two are \textit{key} columns, underlined in Fig.~\ref{fig:overview}.
Using one of our suggested prompting methods (\S\ref{sec:method}), the LLM then generates in response a final results table $T(\hat{R},C)$, as shown in the bottom of the figure.

\section{Prompting LLMs to Generate Tables}
\label{sec:method}

Given a table description and list of target columns $C$, we examine LLMs performance in generating the corresponding table $T(\hat{R},C)$. We focus exclusively on the knowledge stored in the LLM, with retrieval-augmented methods \cite{Lewis2020RetrievalAugmentedGF, yoran2023making} being orthogonal to our study.    

We implement three prompting methods to generate tables, presented in Fig.~\ref{fig:overview}. 
Our first approach (a) prompts the LLM to generate the \emph{full table} all at once. However, as the target table may be quite large (e.g. evaluation tables in \S\ref{sec:dataset} have 1.5K tokens on average), we also experiment with a modular prompting approach \cite{khot2022decomposed}.
Therefore we introduce two additional prompting methods that generate tables in a gradual, (b) \emph{row-by-row} or (c) \emph{cell-by-cell} manner. An in-depth example of our prompting methods is provided in Fig.~\ref{fig:architecture}. Note that all prompts in the figure are appended with the table description and columns (prompt 1 in  Fig.~\ref{fig:architecture}). 
Next, we detail the prompts used for each of our methods. All of our prompts can be found in \S\ref{appendix:prompts} and in our public repository.

\begin{figure*}[ht]
    \centering

    \begin{subfigure}{0.3\textwidth}
        \centering
        \includegraphics[width=\textwidth]{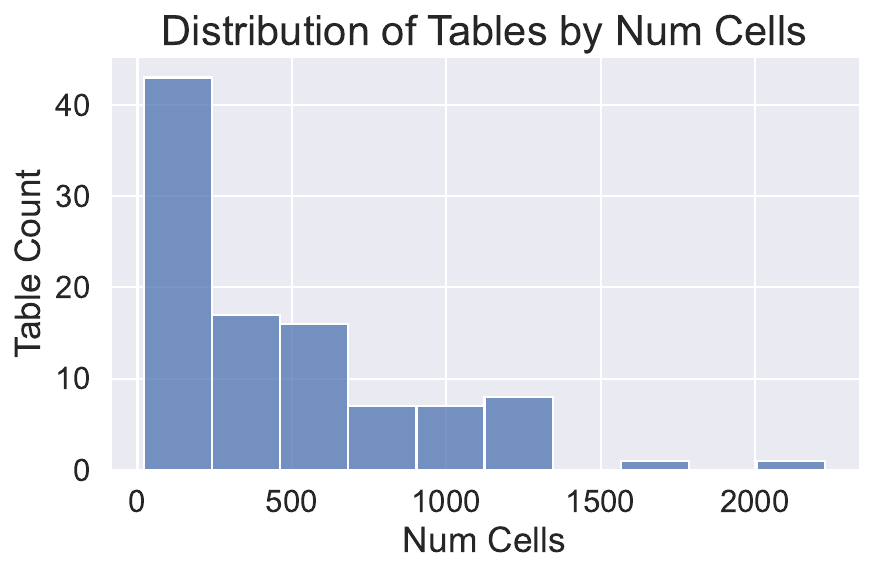}
        \caption{Number of cells per table}
    \end{subfigure}
    \hfill
    \begin{subfigure}{0.3\textwidth}
        \centering
        \includegraphics[width=\textwidth]{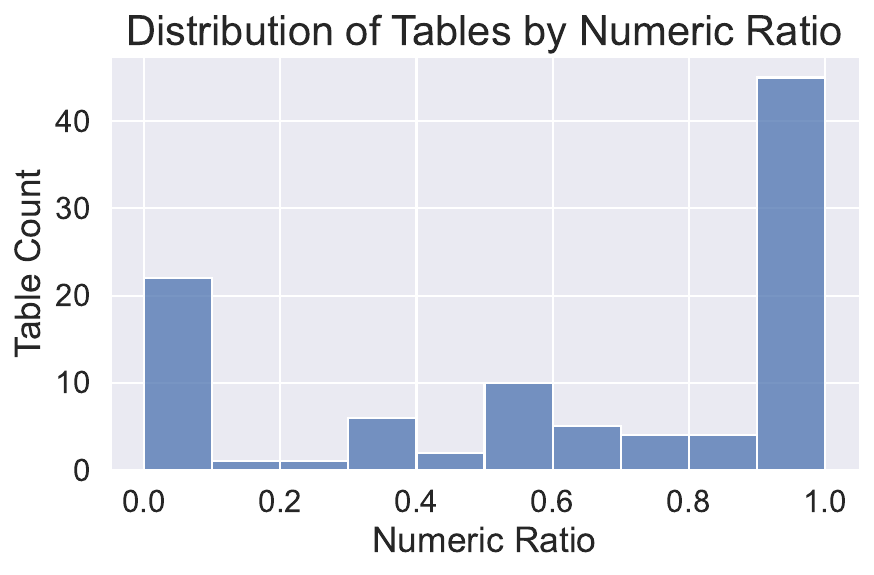}
        \caption{Numeric columns ratio}
    \end{subfigure}
    \hfill
    \begin{subfigure}{0.3\textwidth}
        \centering
        \includegraphics[width=\textwidth]{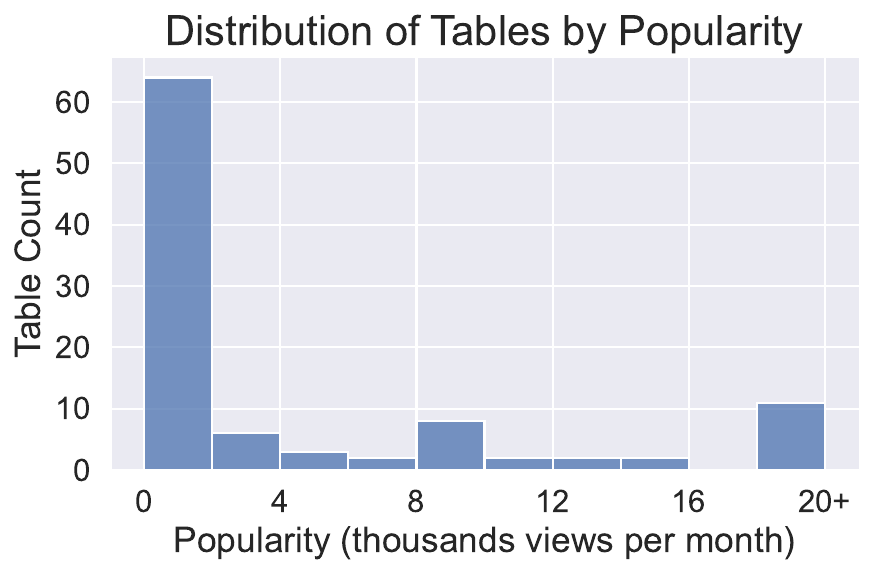}
        \caption{Table popularity}
    \end{subfigure}

    \caption{\bench\  properties distribution: Number of Cells, Ratio of Numeric Columns, and Table Popularity.}
    \label{fig:distributions}
\end{figure*}

\vspace{2mm}
\noindent\textbf{(a) Full-table.} Given the table description and target columns The LLM is prompted to generate all table rows. Example prompts are prompts 1 and 2.A in Fig.~\ref{fig:architecture} which are both concatenated and provided as input to the LLM.

\vspace{2mm}
\noindent\textbf{(b) Row-by-row.} We implement a two-stage prompting method, prompting to separate instances of the LLM. First, we prompt the LLM  for \emph{key generation} to generate the values of all the key columns $C_k$. As key values are a unique identifier for each table entry (\S\ref{sec:problem}), we prompt a second instance of the LLM, to generate the full table row, for a given key value. Thus, for each key cell $\hat{r}_i[C_k]$  that was output by the first LLM, we generate a subsequent prompt to retrieve the remaining row entries $\hat{r}_i[C \setminus C_k]$. 
In total, we generate $|\hat{R}|+1$ prompts, where $|\hat{R}|$ is the number of key values output by the \emph{key generation} prompted LLM. 

In Fig.~\ref{fig:architecture}, box 2.B describes the key generation prompt. Given the table description, and key columns \textit{competition} and \textit{year}, the LLM generates a list of corresponding years and competitions which Susen Tiedtke participated in. Next, each key value returned by the first LLM, is used to generate the remaining row entries. Prompt 3.A prompts the \emph{row generation} LLM to populate columns \texttt{venue}, \texttt{position} which correspond to key $\langle$\textit{``European Junior''}, \textit{``1987''}$\rangle$. The generated values being \textit{``Birmingham''}, and ``3rd''.
A new row-by-row prompt is then generated for the following keys, e.g. $\langle$\textit{``World Championship''}, \textit{``1991''}$\rangle$. 

\vspace{2mm}
\noindent\textbf{(c) Cell-by-cell.} This is a two-stage approach that generates each table cell individually. The first stage is identical to row-by-row, using prompt 2.B to generate all key column values. Then, we use a separate prompt for each table cell, rather than a full row. For each column $c \in C \setminus C_k$ we create a dedicated prompt to generate the cell $\hat{r}_i[c]$, based on the target column and the generated key for $r_i$. 
In total, we use $|\hat{R}|\cdot|C \setminus C_k|+1$ prompts, one to generate the keys, and $|\hat{R}|\cdot|C \setminus C_k|$ to generate each of the non-key cells.

Prompt 3.B in Fig.~\ref{fig:architecture} describes the cell-by-cell method. Given key $\langle$\textit{``European Junior''}, \textit{``1987''}$\rangle$, the corresponding cell in column \texttt{venue} is generated (\textit{Birmingham}). The same prompt is then used for different keys and columns (\texttt{position}).

\vspace{2mm}
\noindent\textbf{Generated Output Format.}
When prompting LLMs we instruct it to return the output in JSON format, also shown in Fig.~\ref{fig:architecture}. We chose JSON following past work \cite{singha2023tabular} and based on our own results. Namely, observed better performance compared to formats such as CSV and SQL, when evaluated on our held-out development set (see \S\ref{sec:dataset}). For the row-by-row and cell-by-cell methods, we process and merge all individual JSON responses to construct the full output table.

\section{\bench\ Benchmark}
\label{sec:dataset}

To evaluate our methods (\S\ref{sec:method}), we introduce a new table generation benchmark called \bench{}. An instance of \bench\ consists of a short manually written description $d$, a list of target columns $C$ and a corresponding table $T=(R,C)$.
As this benchmark targets LLM table generation, based on their parametric knowledge, we considered several key priniciples in its construction:

\begin{itemize}[noitemsep, topsep=0pt, left=0pt]
    \item \emph{Information Coverage}: evaluation tables must contain complete information to avoid instances where the LLM generates accurate entries that are missing from the ground-truth tables.
    \item \emph{Factual Consistency}: tables should include static factual data, to ensure consistent evaluation over time as LLMs evolve \cite{zhang-choi-2021-situatedqa}.
    \item \emph{Conciseness}: table cells should contain concise string, categorical or numeric information, to avoid long descriptive text which is more difficult to evaluate compared to the ground truth.
    \item \emph{Diversity}: the benchmark should include a diverse set of tables in terms of structural properties such as size, data types (e.g., numeric data ratio), and the table ``popularity'' which may indicate the prevalence of its information during the LLM's pre-training \cite{Mallen2022WhenNT}.
\end{itemize}

\begin{figure}[tb]\setlength{\belowcaptionskip}{-8pt}
  \centering
  \includegraphics[clip, width=0.49\textwidth]{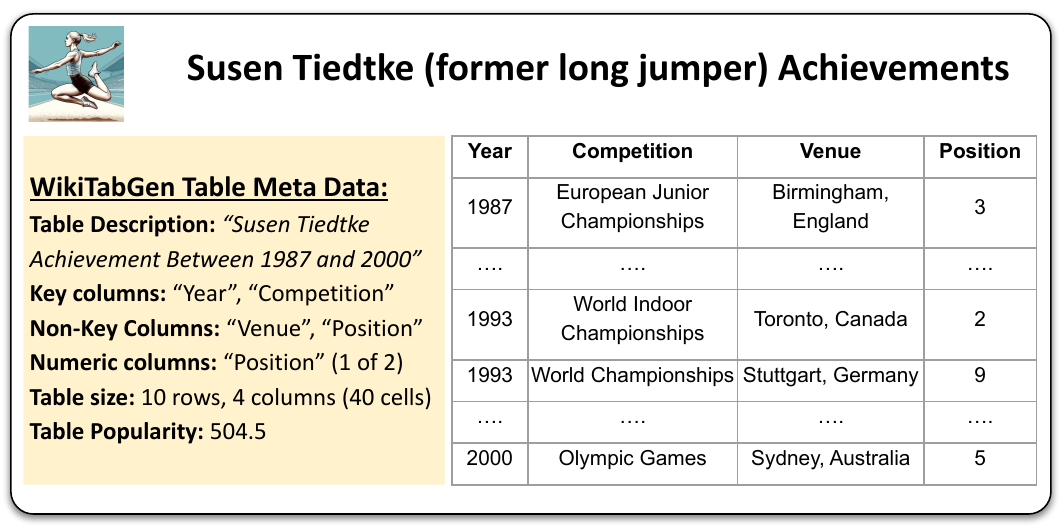}
  \vspace*{-0.7cm}
  \caption{\bench{} example table and meta-data.}
  \label{fig:ex_table}
\end{figure}

Following these principle, opted to use table from Wikipedia, as our evaluation benchmark. Wikipedia information is normally used to evaluate the closed-book performance of LLMs, as it contains factual and objective information \cite{kwiatkowski-etal-2019-natural}. This contrasts certain tabular datasets which may contain domain-specific data \cite{Yu2018SpiderAL}. Furthermore, as Wikipedia data forms part of LLMs pre-training data \cite{NEURIPS2020_1457c0d6,Touvron2023LLaMAOA}, we view it as ideal to gauge the extent to which such models memorize tabular data.

To build our benchmark, we iterate over the Wikipedia tables provided by \citet{Bhagavatula2015TabELEL}.\footnote{Creative Commons Attribution 4.0 International License.}
First, we discard all tables that are ``non-relational'', i.e. contain a composite header, nested tables or inverted tables. We additionally remove tables that are too small, where $|R|<10$ or $|C|<2$. 
Next, we manually selected 119 random tables with diverse number of columns, rows and portion of numeric values (numbers and dates). To ensure evaluation coverage we removed columns with partial entries. In addition, columns containing long texts were omitted to ensure a concise evaluation. 
Lastly, we manually annotated each table with a short description in natural language to describe its contents. We chose to manually write table descriptions as the original table captions were often ambiguous or did not accurately describe the table. Furthermore, as Wikipedia tables may expand over time (e.g. new NBA championship teams), we made sure to specify the concrete time-frame in such examples, following \citet{zhang-choi-2021-situatedqa}, e.g.  ``\textit{George Clooney Films \underline{released between 1983 and 2013}}''

As shown in Fig.~\ref{fig:ex_table}, each table in \bench\ is provided with additional metadata, consisting of its: text description; table size (number of columns, rows and cells); key-columns; numeric columns (containing numbers or dates); and table popularity. Inspired by \citet{Mallen2022WhenNT}, we define \emph{table popularity} as the average number of monthly views to the Wikipedia page containing the said table. To measure pages views we use the Wikipedia API.\footnote{\url{https://api.wikimedia.org}}

Overall, \bench{} has 119 examples, where 100 tables are used for evaluation (\S\ref{sec:experiments}) and the remaining 19 are a held-out development set which we used when implementing our methods.
In Fig.~\ref{fig:distributions} we plot the distribution of three of the aforementioned properties in \bench{}: size, numeric column ratio and popularity. In terms of size, our evaluation tables have 77.5 rows, 6.9 columns and 453 cells on average, with an average length of 1,497 tokens. The average percentage of numeric columns per table is 62\% of columns, showcasing the prevalence of numerical data in our tables. In terms of table popularity, the average number of monthly views per table is 8,449. In \S\ref{sec:analysis} we further explore each of these properties and evaluate their effect on table generation performance. 



\section{Experimental Setting}
\label{sec:experiments}
We describe our experimental setting for evaluating the table generation capabilities of LLMs. All models were evaluated on the \bench\ benchmark. Next, we list the LLMs, prompts and evaluation methods used for table generation. Last, we detail our different experimental scenarios.

\subsection{Language Models and Prompts}
In our experiments we evaluate four popular LLMs. The first two are the state-of-the-art mdels from OpenAI \cite{Achiam2023GPT4TR}: GPT-3.5 (\texttt{gpt-3.5-turbo-instruct-0914}) and GPT-4 (\texttt{gpt-4-1106-preview}). In addition, we evaluate two open-source LLMs from MetaAI: \cite{Touvron2023Llama2O}: Llama2-13B (\texttt{Llama-2-13b-chat-hf}) and Llama2-70B (\texttt{Llama-2-70b-chat-hf}).
The same prompts described in \S\ref{sec:method} were used across all four LLMs, with the generation temperature set to zero.
When developing our prompting methods, we used the held-out set of 19 tables, described in \S\ref{sec:dataset}. This set was reserved for prompt engineering (experimenting with different output formats JSON, CSV, etc.) and is not included in the evaluation set of \bench{}.


\begin{table*}[htb]
    \centering
    \scriptsize
    \begin{adjustbox}{width=\textwidth}
    \begin{tabular}{llccccccccc}
        \toprule
        \multirow{2}{*}{\textbf{LLM}} & \multirow{2}{*}{\textbf{Method}} & \multicolumn{3}{c}{\textbf{Keys}} & \multicolumn{3}{c}{\textbf{Non-Keys}} & \multicolumn{3}{c}{\textbf{Overall}} \\
        \cmidrule(lr){3-5} \cmidrule(lr){6-8} \cmidrule(lr){9-11}
        & & Recall & Precision & F1 & Recall & Precision & F1 &  Recall & Precision & F1 \\
        \midrule
        \multirow{2}{*}{GPT-3.5} & Full tab. & 44.6\% & 59.4\% & 46.4\% & 9.0\% & 13.3\% & 9.6\% &  15.4\% & 21.2\% & 16.1\% \\
        & Row-by-row & 51.5\% & 55.6\% & 49.4\% & 7.0\% & 10.3\% & 7.2\% &  14.6\% & 18.0\% & 14.3\% \\
        & Cell-by-cell & 51.5\% & 55.6\% & 49.4\% & 7.5\% & 10.4\% & 7.6\% & 15.0\% & 18.1\% & 14.6\% \\
        \midrule
        \multirow{2}{*}{GPT-4} & Full tab. & 40.6\% & 62.5\% & 43.8\% & 10.4\% & 18.0\% & 11.5\% &  16.0\% & 25.7\% & 17.5\% \\
        & Row-by-row & 55.8\% & 55.6\% & \bf 53.7\% & 12.3\% & 13.5\% & \bf 12.2\% &  20.0\% & 21.1\% & \bf 19.6\% \\
        & Cell-by-cell & 55.7\% & 55.6\% & \bf 53.7\% & 11.2\% & 12.5\% & 11.1\% & 19.0\% & 20.1\% & 18.6\% \\
        \midrule
        \multirow{2}{*}{Llama2-13B} & Full tab. & 23.8\% & 50.5\% & 28.1\% & ~2.0\% & ~6.2\% & ~2.6\% &  ~6.2\% & 13.8\% & ~7.4\% \\
        & Row-by-row & 31.8\% & 43.0\% & 32.4\% & ~2.3\% & ~4.6\% & ~2.6\%  & ~7.2\% & 10.8\% & ~7.5\% \\
        & Cell-by-cell & 31.9\% & 43.0\% & 32.5\% & ~0.8\% & ~1.5\% & ~0.8\% & ~6.0\% & ~8.7\% & ~6.2\% \\
        \midrule
        \multirow{2}{*}{Llama2-70B} & Full tab. & 24.7\% & 49.0\% & 29.0\% & ~2.5\% & ~7.4\% & ~3.4\% & ~6.7\% & 14.6\% & ~8.0\% \\
        & Row-by-row & 42.1\% & 46.6\% & 41.4\% & ~3.3\% & ~5.5\% & ~3.4\% & ~9.3\% & 12.1\% & ~9.3\% \\
        & Cell-by-cell & 42.1\% & 46.6\% & 41.4\% & ~2.3\% & ~4.4\% & ~2.4\% & ~8.5\% & 11.1\% & ~8.4\% \\
        \bottomrule
    \end{tabular}
    \end{adjustbox}
        \caption{Table generation performance metrics for the different models and prompting methods.}
    \label{tab:results_main}
\end{table*}


\subsection{Evaluation Method} 
\label{ssec:evaluation_method}
We evaluate the accuracy of LLM generated tables against the ground-truth tables. Unlike list QA \cite{Amouyal2022QAMPARIA}, when evaluating table generation we compare two list of \emph{table rows}, where each row has multiple cells. We use a two-step evaluation which first aligns rows based on their key values, then matches each cell in the row.

Given output table $\hat{T}(\hat{R}, C)$ and ground-truth table $T(R, C)$, we first align the rows $\hat{R}$ to their corresponding rows in $R$ by matching their respective keys, namely $\hat{r} \leadsto r \iff \hat{r}[C_k] = r[C_k]$. For rows with multiple key columns, all values must be identical. We check for an \emph{exact match} of the cell content and allow a $\pm 0.1\%$ error for numeric values (in \S\ref{app:fuzzy}, we describe how we compare date values and handle null, missing and duplicate cells).
A \textit{correct} cell in $T(\hat{R}, C)$ is a cell $\hat{r}[c]$ such that $\hat{r} \leadsto r \wedge \hat{r}[c] = r[c]$ i.e. row $\hat{r}$ is aligned with $r$ in the ground-truth table, and their corresponding values in column $c$ are identical.
We then calculate \emph{Table Precision} as $\frac{\text{\# Correct Cells}}{\text{\# Generated Cells}}$, and \emph{Table Recall} as $\frac{\text{\# Correct Cells}}{\text{\# Ground-Truth Cells}}$ and corresponding F1 score.

For our analysis in \S\ref{sec:analysis}, we also consider the precision, recall, and F1 scores separately for \textit{keys} and \textit{non-keys}. The \textit{keys} scores are calculated based on the number of matching keys, where for each row all the cells of $C_k$ must match. For non-key cell scores we consider only cells in $C \setminus C_k$. We provide the formulas in Appendix~\ref{app:fuzzy}.

\subsection{Table Generation Scenarios}
In addition to the table generation scenario described in \S\ref{sec:problem}, where the generation request contains only the table description and list of columns, we considered two alternative scenarios where additional information is provided to the LLM:

\textbf{Table Row Example.} In this scenario, in addition to the description and list of columns, we also provide the LLM with an example row $r[C]$ from the target table. We examine if such an example improves the LLM's performance in generating the rest of the table. We tested this scenario on all prompting methods (\S\ref{sec:method}) by concatenating the first row of the target table to the table description.

\textbf{Oracle Keys.} This ablation provides the LLM the ground-truth set of keys cells $R[C_k]$ and measures the model's performance in generating the remaining cells. This scenario is particularly relevant for applications where the keys are known in advance, and the task involves filling in the associated data. We conducted this experiment for both the row-by-row and cell-by-cell prompting methods by skipping the keys generation prompt (prompt 2.B in Fig.~\ref{fig:architecture}), and providing the ground-truth keys instead.

\section{Results and Analysis } 
\label{sec:analysis}
We begin by describing our main results for table generation, for each of our LLMs and prompting methods. Then, we present the results for our additional scenarios, when including the first table row or the ground-truth of key values. Next, we analyze the generation cost and accuracy trade-offs of each prompting methods. Lastly, we examine the effect of table properties (size, numeric data, and table popularity) on the LLM generation performance.

\subsection{Table Generation Performance}

Tab.~\ref{tab:results_main} presents the table generation results of our three prompting methods, when evaluated for each LLM on \bench{}. We list the precision, recall, and F1 scores for keys, non-keys, and the full tables (averaged across all tables). GPT-4 is the top-performing model, consistently surpassing the other LLMs, and reaches the highest overall F1 score, when using the row-by-row prompt (19.6\%).

For all LLMs, we observe that the row-by-row and cell-by-cell methods significantly improve the \emph{keys} generation performance (see keys F1 scores in Tab.~\ref{tab:results_main}). This demonstrates the effectiveness of separating the table generation into keys and non-key cells. Besides GPT-3.5, for GPT-4, Llama2-13B, and Llama2-70B, the top performance was using the row-by-row method. We note that the keys generation performance is significantly higher that for non-key cells (53.7\% compared to 12.5\%- 19.5\% for GPT-4).

\begin{table*}[htb]
\centering
\scriptsize
\begin{adjustbox}{width=0.90\textwidth}
\begin{tabular}{llccccccccc}
\toprule
        \multirow{2}{*}{\textbf{LLM}} & \multirow{2}{*}{\textbf{Method}} & \multicolumn{2}{c}{\textbf{Keys F1 (\%)}} & \multicolumn{2}{c}{\textbf{Non-Keys F1 (\%)}} & \multicolumn{2}{c}{\textbf{Overall F1 (\%)}} \\
        \cmidrule(lr){3-4} \cmidrule(lr){5-6} \cmidrule(lr){7-8}
        & & No-Example & Example & No-Example & Example & No-Example & Example\\
\midrule
\multirow{2}{*}{GPT-3.5} & Full tab. & 46.0 & \textbf{49.0} & ~9.4 & \textbf{10.9} & 15.9 & \textbf{17.9} \\
& Row-by-row & \textbf{49.6} & 49.1 & ~7.0 & ~\textbf{7.6} & 14.2 & \textbf{14.6}\\
\midrule
\multirow{2}{*}{GPT-4} & Full tab. & 43.4 & \textbf{49.4} &  11.3 & \textbf{14.5} & 17.2 & \textbf{21.4}\\
 & Row-by-row & 55.1 & \textbf{55.3} &  12.5 & \textbf{14.8} & 20.1 & \textbf{21.9}\\
\midrule
\multirow{2}{*}{Llama2-13B} & Full tab. & 27.2 & \textbf{27.5} & ~2.3 & ~\textbf{3.9} & ~7.0 & ~\textbf{8.4}\\
& Row-by-row & \textbf{32.5} & 32.4 & ~2.6 & ~\textbf{3.5} & ~7.5 & ~\textbf{8.4}\\
\midrule
\multirow{2}{*}{Llama2-70B} & Full tab. & 28.4 & \textbf{31.3} & ~3.2 & ~\textbf{5.0} & ~7.8 & ~\textbf{9.7}\\
& Row-by-row & \textbf{39.8} & 39.7 & ~3.1 & \textbf{3.8} & ~8.9 & ~\textbf{9.5}\\
\bottomrule
\end{tabular}
\end{adjustbox}
\caption{Performance comparison with and without an example row, using full table and row-by-row methods. 
}
\label{tab:results_first_row}
\end{table*}

\subsection{Additional Generation Scenarios}
We measure the effect of providing additional information during table generation: (1) an example row, (2) the ground-truth table keys.

\textbf{Table Row Example.} Tab.~\ref{tab:results_first_row} lists the performance results when including an example row from the target table (as we omit the example row from the F1 calculations our results slightly differ from Tab.~\ref{tab:results_main}). We chose to omit cell-by-cell evaluation, as row-by-row consistently outperformed the approach and due to the high cost of cell-by-cell (\S\ref{ssec:costs}).
The overall F1 score is consistently higher when an example row is added. As shown in the table, this improvement generally stems from better performance in the \textit{non-keys} generation rather than in generating the table keys.

\begin{table}[tb]
\centering
\scriptsize
\begin{adjustbox}{width=0.5\textwidth}
\begin{tabular}{lcccc}
\toprule
        \multirow{2}{*}{\textbf{LLM}} &  \multicolumn{2}{c}{\textbf{Non-Keys F1 (\%)}} & \multicolumn{2}{c}{\textbf{Overall F1 (\%)}} \\
        \cmidrule(lr){2-3} \cmidrule(lr){4-5} 
        & Base. & Orac. & Base. & Orac. \\
\midrule
GPT-3.5 &  ~7.2 & 17.6 (\underline{+10.4}) & 14.3 & 35.0 (\underline{+20.7}) \\

GPT-4 &   12.2 & 23.9  (\underline{+11.7}) & 19.6 & 40.0 (\underline{+20.4})\\

Llama2-13B &  ~2.6 & 10.3 (\underline{+7.7}) & ~7.5 & 29.5 (\underline{+22.0})\\

Llama2-70B &  ~3.4 & 12.8 (\underline{+9.4}) & ~9.3 & 31.4 (\underline{+22.1})\\
\bottomrule

\end{tabular}
\end{adjustbox}
\caption{Performance comparison of the row-by-row method with and without oracle keys.}
\label{tab:results_oracle}
\end{table}

\textbf{Oracle Keys.} Tab.~\ref{tab:results_oracle} describes the performance of all LLMs, using the row-by-row method, when given the ground-truth key values.
As expected, the overall F1 scores are significantly higher when using oracle keys, because now $\hat{R}[C_k] = R[C_k]$. We observe an additional improvement in the non-keys F1, which is expected as more table rows were aligned to the target table (given the keys), and thus more cells were successfully matched.

\subsection{Prompting Cost Analysis}
\label{ssec:costs}
We analyze the performance of our prompting methods as a function of their overall cost.

\begin{figure}[t]
  \includegraphics[width=0.45\textwidth]{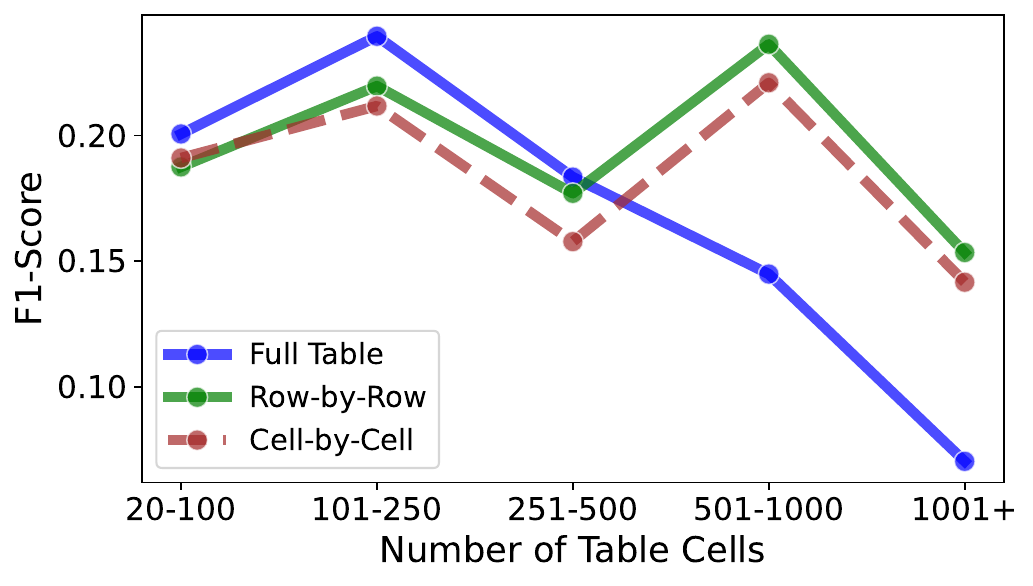}
  \caption{Prompting performance w.r.t. table size.}
  \label{fig:prompting_performance}
\end{figure}

Our motivation for both the row-by-row and cell-by-cell methods was to generate larger tables, by breaking down table generation into two stages.
Therefore, in Fig.~\ref{fig:prompting_performance} compare the performance of out three prompting methods as a function of table size. We measure the F1 scores on tables between 20 to 1,000+ cells. For smaller tables, with up to 500 cells, all prompting methods perform roughly the same. However, as the number of cells increases, both the row-by-row and cell-by-cell methods outperform the full table generation.

\begin{figure}[t]
  \centering
  \includegraphics[clip, width=0.44\textwidth]{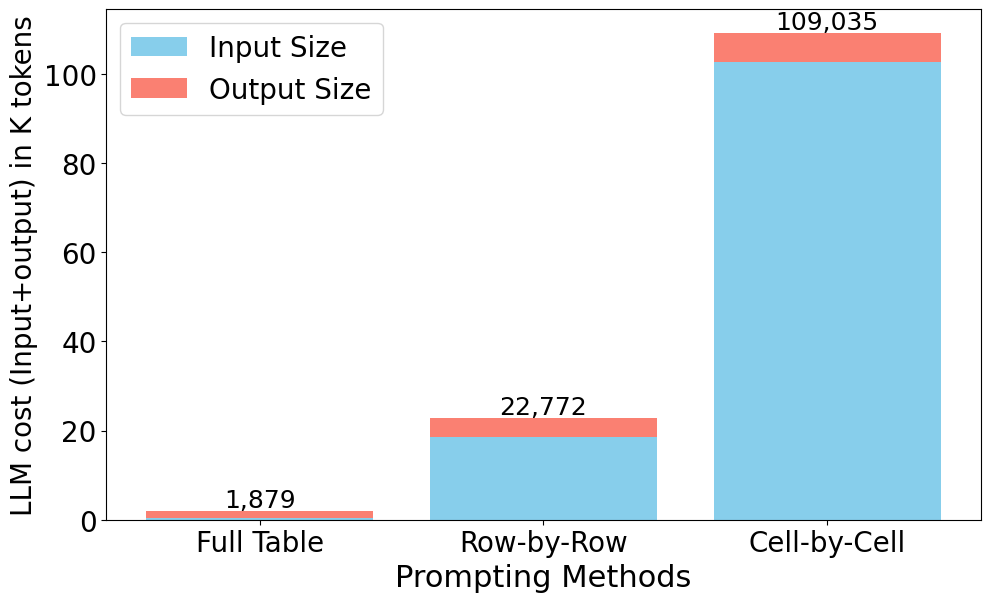}
  \vspace*{-0.3cm}
  \caption{Cost analysis of prompting methods.}
  \label{fig:cost}
\end{figure}

However, while outperforming full-table, as table size increase, both row-by-row and cell-by-cell require significantly more tokens in their prompts, shown in Fig.~\ref{fig:cost}. Therefore, these methods incur much higher costs, in particular when employing commercial LLMs.\footnote{Costs are calculated based on GPT-4 prompt sizes, with similar results also obtained for the other LLMs.} The output number of tokens is roughly similar for all approaches however, the two-stage approach have a significantly larger input due to the repeated use of distinct row and cell generation prompts (prompts 3.A, 3.B in Fig.~\ref{fig:architecture}).



\begin{figure*}[tb]
    \centering
    \begin{subfigure}{0.32\textwidth}
        \centering
        \includegraphics[width=\linewidth]{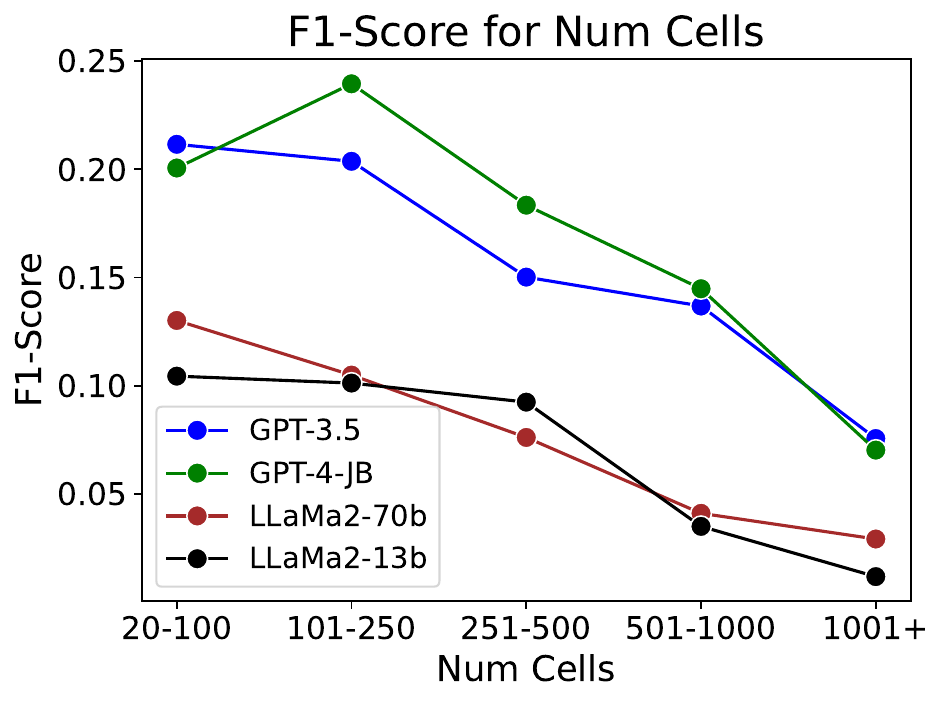}
        \caption{Table Cells Num. - Table F1}
        \label{sub:numcells}
    \end{subfigure}
    \hfill
    \begin{subfigure}{0.32\textwidth}
        \centering
        \includegraphics[width=\linewidth]{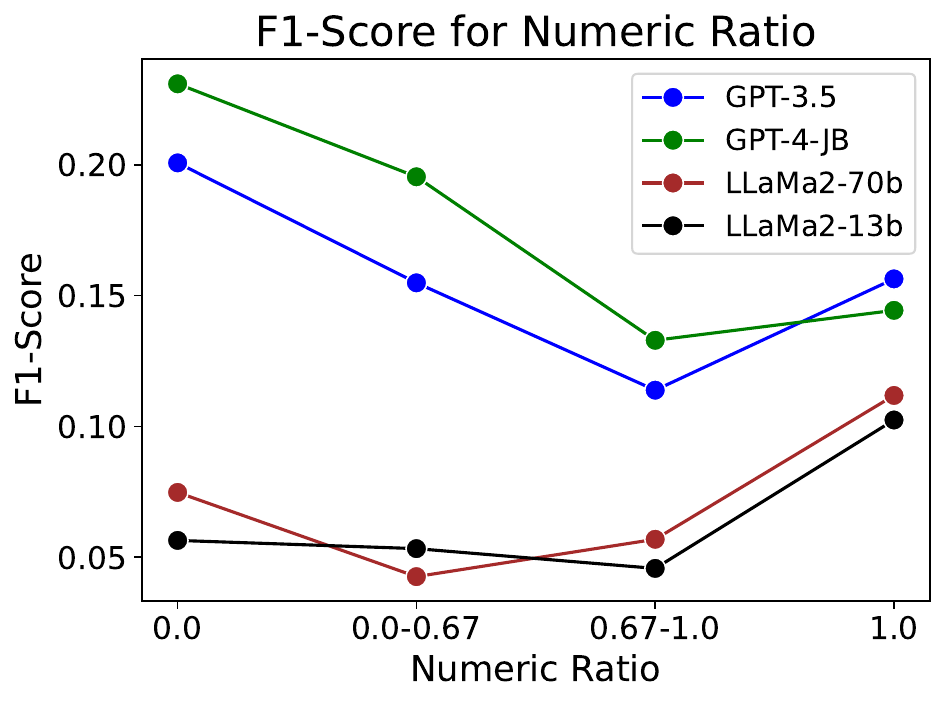}
        \caption{Numeric Cols. \% - Table F1}
        \label{sub:numeric_cols}
    \end{subfigure}
    \hfill
    \begin{subfigure}{0.32\textwidth}
        \centering
        \includegraphics[width=\linewidth]{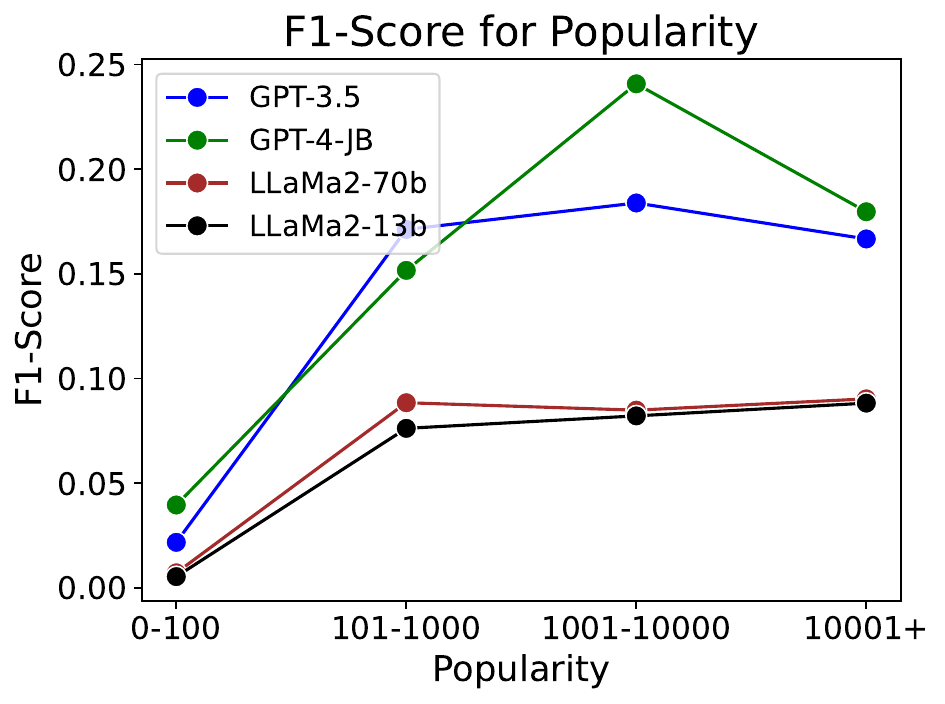}
        \caption{Table Popularity - Table F1}
        \label{sub:popularity}
    \end{subfigure}
    \caption{The effect of table size, the ratio of numeric columns, and table popularity on the generation performance of the full-table method, with four different LLMs.  }
    \label{fig:comparison}
\end{figure*}

\subsection{Table Properties Effect on Performance}
\label{ssec:properties_effects}
As noted in \S\ref{sec:dataset}, we systematically measure the effect of table properties such as the size, numeric data and table popularity affect the LLM generation performance.

Fig.~\ref{fig:comparison} displays the table F1 scores as a function of the number of table cells, percentage of numerical data columns (number or date cells) and the table popularity score. These results are provided for all four of experiments LLMs, using the full-table generation method. As our aim is to measure the effect each property has on the LLM (not to compare different methods), we consider this to be an apt choice of setting. A further breakdown of the properties' effect on the keys and non-keys F1 scores is provided in \S\ref{app:comparison}.

As shown in Fig.~\ref{sub:numcells}, the larger the table, the lower the F1 scores are for all LLMs. In \S\ref{ssec:costs} we observed this trend to be less apparent for the row-by-row and cell-by-cell methods.

Fig.~\ref{sub:numeric_cols} measures the effect of that the percentage of columns containing numbers or dates has on performance. We observe a decrease in F1 as the portion of numerical content is higher. We attribute the slight increase in fully numeric to the fact that such tables in \bench\ contain demographic information where the key column is \textit{year}, which is easier for the LLMs to generate.

Fig.~\ref{sub:popularity} displays the positive effect of table popularity on performance. This potentially stems from the prevalence of more popular Wikipedia pages (or related entities) in the LLMs' training data. Unsurprising, the less common the tabular information is, the more difficult it is for the LLM to generate.

From this analysis, we conclude that generating tables from LLMs' parametric knowledge is more challenging when the tables are larger, have a high portion of numerical data, or concern less topics.

\section{Related Work}
\label{sec:related_work}
Machine reasoning on table using pre-trained LLMs has largely been explored in the context of data augmentation \cite{Borisov2022LanguageMA, Zhang2023GenerativeTP} to improve the performance on downstream tasks. The focus has largely been on tasks where a table is provided as input to the model namely: QA over tables \cite{Chen2020HybridQAAD, chen-etal-2022-convfinqa, Seedat2023CuratedLS}, text-to-SQL translation \cite{Deng2021StructureGroundedPF, wolfson-etal-2022-weakly}, table editing \cite{Li2023TableGPTTG, Sui2023TableML} and table-to-text generation \cite{Parikh2020ToTToAC}. Conversely, our approach receives only a user query and schema as input, and is tasked with generating an entire table.

Closest to ours are the recent table generation datasets by \citet{Pal2023MultiTabQAGT, Akhtar2024TANQAO}. In both works the LLM is provided with a user query (in text or SQL) and is tasked with generating a table, as the query answer. \citet{Pal2023MultiTabQAGT} evaluate on tables from the Spider dataset \cite{Yu2018SpiderAL}, which contains domain-specific information that is less likely to be stored in the parametric knowledge of LLMs. Similar to us, \citet{Akhtar2024TANQAO} rely on Wikipedia however, they automatically construct new tables which are relatively small (average of 6.7 rows, 4 columns). By comparison our evaluation is on larger tables with the median number of rows being 48 (average of 77.5 rows, 6.9 columns). This emphasizes our focus on extracting long-form tabular data from LLMs, thereby extending past attempts on KBs and text \cite{Cohen2023CrawlingTI, Mallen2022WhenNT, Carlini2022QuantifyingMA}.

Our key generation phase in \S\ref{sec:method} is an instance of a list question answering problem. The challenge of list QA in LLMs has been explored in recent works \cite{Amouyal2022QAMPARIA,  Malaviya2023QUESTAR}. However, we further expand this challenge by focusing on generating the entire table.

\section{Conclusion}
This paper explores the capability of state-of-the-art LLMs to generate entire tables, by relying exclusively on their parametric knowledge.
We introduced three prompt-based table generation methods and evaluated them on our newly constructed benchmark, \bench\. Our results and analysis of four state-of-the-art models underscore the overall challenge which table generation poses to LLMs. We hope that \bench\ and our comprehensive analysis will provide a concrete framework for future research on table generation using LLMs.

\section{Limitations}
\label{sec:limitations}

We now list the limitations to our work. 

Our first limitation is the size of the \bench\ evaluation benchmark, which contains 100 tables. We atribute this size mainly to the high generation costs of running state-of-the-art LLMs on large tables \S\ref{ssec:costs}. As noted in \S\ref{sec:dataset}, the tables in \bench\ contain close to 1,500 tokens on average, evaluating them using commercial, state-of-the-art LLMs is non-trivial. 

Second, all tables in \bench\ are tied to a single source, i.e. Wikipedia articles. This choice was made to ensure that the underlying information exists in common LLMs training data. However, we did not examine the performance on tables generated from other sources, such as news articles or tables that require multi-source integration.

Third, our evaluation metric for generation performance is based on a strict, almost exact comparison of cell values (see \S\ref{ssec:evaluation_method} and \S\ref{app:fuzzy}). This approach can sometimes result in LLMs being penalized for minor formatting errors. However, we chose this rigorous method over more relaxed alternatives (such as those based on semantic similarity or fuzzy matching) because the \bench\ table cells contain concise categorical or numeric information, often consisting of fewer than four tokens. Consequently, even minor inaccuracies can significantly impact the performance assessment.


\bibliography{anthology,custom}

\begin{thebibliography}{46}
\expandafter\ifx\csname natexlab\endcsname\relax\def\natexlab#1{#1}\fi

\bibitem[{Achiam et~al.(2023)Achiam, Adler, Agarwal, Ahmad, Akkaya, Aleman, Almeida, Altenschmidt, Altman, Anadkat, Avila, Babuschkin, Balaji, Balcom, Baltescu, Bao, Bavarian, Belgum, Bello, Berdine, Bernadett-Shapiro, Berner, Bogdonoff, Boiko, Boyd, Brakman, Brockman, Brooks, Brundage, Button, Cai, Campbell, Cann, Carey, Carlson, Carmichael, Chan, Chang, Chantzis, Chen, Chen, Chen, Chen, Chen, Chess, Cho, Chu, Chung, Cummings, Currier, Dai, Decareaux, Degry, Deutsch, Deville, Dhar, Dohan, Dowling, Dunning, Ecoffet, Eleti, Eloundou, Farhi, Fedus, Felix, Fishman, Forte, Fulford, Gao, Georges, Gibson, Goel, Gogineni, Goh, Gontijo-Lopes, Gordon, Grafstein, Gray, Greene, Gross, Gu, Guo, Hallacy, Han, Harris, He, Heaton, Heidecke, Hesse, Hickey, Hickey, Hoeschele, Houghton, Hsu, Hu, Hu, Huizinga, Jain, Jain, Jang, Jiang, Jiang, Jin, Jin, Jomoto, Jonn, Jun, Kaftan, Kaiser, Kamali, Kanitscheider, Keskar, Khan, Kilpatrick, Kim, Kim, Kim, Kirchner, Kiros, Knight, Kokotajlo, Kondraciuk, Kondrich, Konstantinidis, Kosic,
  Krueger, Kuo, Lampe, Lan, Lee, Leike, Leung, Levy, Li, Lim, Lin, Lin, Litwin, Lopez, Lowe, Lue, Makanju, Malfacini, Manning, Markov, Markovski, Martin, Mayer, Mayne, McGrew, McKinney, McLeavey, McMillan, McNeil, Medina, Mehta, Menick, Metz, Mishchenko, Mishkin, Monaco, Morikawa, Mossing, Mu, Murati, Murk, M'ely, Nair, Nakano, Nayak, Neelakantan, Ngo, Noh, Long, O'Keefe, Pachocki, Paino, Palermo, Pantuliano, Parascandolo, Parish, Parparita, Passos, Pavlov, Peng, Perelman, de~Avila Belbute~Peres, Petrov, de~Oliveira~Pinto, Pokorny, Pokrass, Pong, Powell, Power, Power, Proehl, Puri, Radford, Rae, Ramesh, Raymond, Real, Rimbach, Ross, Rotsted, Roussez, Ryder, Saltarelli, Sanders, Santurkar, Sastry, Schmidt, Schnurr, Schulman, Selsam, Sheppard, Sherbakov, Shieh, Shoker, Shyam, Sidor, Sigler, Simens, Sitkin, Slama, Sohl, Sokolowsky, Song, Staudacher, Such, Summers, Sutskever, Tang, Tezak, Thompson, Tillet, Tootoonchian, Tseng, Tuggle, Turley, Tworek, Uribe, Vallone, Vijayvergiya, Voss, Wainwright, Wang, Wang,
  Wang, Ward, Wei, Weinmann, Welihinda, Welinder, Weng, Weng, Wiethoff, Willner, Winter, Wolrich, Wong, Workman, Wu, Wu, Wu, Xiao, Xu, Yoo, Yu, Yuan, Zaremba, Zellers, Zhang, Zhang, Zhao, Zheng, Zhuang, Zhuk, and Zoph}]{Achiam2023GPT4TR}
OpenAI~Josh Achiam, Steven Adler, Sandhini Agarwal, Lama Ahmad, Ilge Akkaya, Florencia~Leoni Aleman, Diogo Almeida, Janko Altenschmidt, Sam Altman, Shyamal Anadkat, Red Avila, Igor Babuschkin, Suchir Balaji, Valerie Balcom, Paul Baltescu, Haiming Bao, Mo~Bavarian, Jeff Belgum, Irwan Bello, Jake Berdine, Gabriel Bernadett-Shapiro, Christopher Berner, Lenny Bogdonoff, Oleg Boiko, Madelaine Boyd, Anna-Luisa Brakman, Greg Brockman, Tim Brooks, Miles Brundage, Kevin Button, Trevor Cai, Rosie Campbell, Andrew Cann, Brittany Carey, Chelsea Carlson, Rory Carmichael, Brooke Chan, Che Chang, Fotis Chantzis, Derek Chen, Sully Chen, Ruby Chen, Jason Chen, Mark Chen, Benjamin Chess, Chester Cho, Casey Chu, Hyung~Won Chung, Dave Cummings, Jeremiah Currier, Yunxing Dai, Cory Decareaux, Thomas Degry, Noah Deutsch, Damien Deville, Arka Dhar, David Dohan, Steve Dowling, Sheila Dunning, Adrien Ecoffet, Atty Eleti, Tyna Eloundou, David Farhi, Liam Fedus, Niko Felix, Sim'on~Posada Fishman, Juston Forte, Isabella Fulford, Leo Gao,
  Elie Georges, Christian Gibson, Vik Goel, Tarun Gogineni, Gabriel Goh, Raphael Gontijo-Lopes, Jonathan Gordon, Morgan Grafstein, Scott Gray, Ryan Greene, Joshua Gross, Shixiang~Shane Gu, Yufei Guo, Chris Hallacy, Jesse Han, Jeff Harris, Yuchen He, Mike Heaton, Johannes Heidecke, Chris Hesse, Alan Hickey, Wade Hickey, Peter Hoeschele, Brandon Houghton, Kenny Hsu, Shengli Hu, Xin Hu, Joost Huizinga, Shantanu Jain, Shawn Jain, Joanne Jang, Angela Jiang, Roger Jiang, Haozhun Jin, Denny Jin, Shino Jomoto, Billie Jonn, Heewoo Jun, Tomer Kaftan, Lukasz Kaiser, Ali Kamali, Ingmar Kanitscheider, Nitish~Shirish Keskar, Tabarak Khan, Logan Kilpatrick, Jong~Wook Kim, Christina Kim, Yongjik Kim, Hendrik Kirchner, Jamie~Ryan Kiros, Matthew Knight, Daniel Kokotajlo, Lukasz Kondraciuk, Andrew Kondrich, Aris Konstantinidis, Kyle Kosic, Gretchen Krueger, Vishal Kuo, Michael Lampe, Ikai Lan, Teddy Lee, Jan Leike, Jade Leung, Daniel Levy, Chak~Ming Li, Rachel Lim, Molly Lin, Stephanie Lin, Mateusz Litwin, Theresa Lopez, Ryan
  Lowe, Patricia Lue, Anna~Adeola Makanju, Kim Malfacini, Sam Manning, Todor Markov, Yaniv Markovski, Bianca Martin, Katie Mayer, Andrew Mayne, Bob McGrew, Scott~Mayer McKinney, Christine McLeavey, Paul McMillan, Jake McNeil, David Medina, Aalok Mehta, Jacob Menick, Luke Metz, Andrey Mishchenko, Pamela Mishkin, Vinnie Monaco, Evan Morikawa, Daniel~P. Mossing, Tong Mu, Mira Murati, Oleg Murk, David M'ely, Ashvin Nair, Reiichiro Nakano, Rajeev Nayak, Arvind Neelakantan, Richard Ngo, Hyeonwoo Noh, Ouyang Long, Cullen O'Keefe, Jakub~W. Pachocki, Alex Paino, Joe Palermo, Ashley Pantuliano, Giambattista Parascandolo, Joel Parish, Emy Parparita, Alexandre Passos, Mikhail Pavlov, Andrew Peng, Adam Perelman, Filipe de~Avila Belbute~Peres, Michael Petrov, Henrique~Pond{\'e} de~Oliveira~Pinto, Michael Pokorny, Michelle Pokrass, Vitchyr~H. Pong, Tolly Powell, Alethea Power, Boris Power, Elizabeth Proehl, Raul Puri, Alec Radford, Jack Rae, Aditya Ramesh, Cameron Raymond, Francis Real, Kendra Rimbach, Carl Ross, Bob
  Rotsted, Henri Roussez, Nick Ryder, Mario~D. Saltarelli, Ted Sanders, Shibani Santurkar, Girish Sastry, Heather Schmidt, David Schnurr, John Schulman, Daniel Selsam, Kyla Sheppard, Toki Sherbakov, Jessica Shieh, Sarah Shoker, Pranav Shyam, Szymon Sidor, Eric Sigler, Maddie Simens, Jordan Sitkin, Katarina Slama, Ian Sohl, Benjamin~D. Sokolowsky, Yang Song, Natalie Staudacher, Felipe~Petroski Such, Natalie Summers, Ilya Sutskever, Jie Tang, Nikolas~A. Tezak, Madeleine Thompson, Phil Tillet, Amin Tootoonchian, Elizabeth Tseng, Preston Tuggle, Nick Turley, Jerry Tworek, Juan Felipe~Cer'on Uribe, Andrea Vallone, Arun Vijayvergiya, Chelsea Voss, Carroll~L. Wainwright, Justin~Jay Wang, Alvin Wang, Ben Wang, Jonathan Ward, Jason Wei, CJ~Weinmann, Akila Welihinda, Peter Welinder, Jiayi Weng, Lilian Weng, Matt Wiethoff, Dave Willner, Clemens Winter, Samuel Wolrich, Hannah Wong, Lauren Workman, Sherwin Wu, Jeff Wu, Michael Wu, Kai Xiao, Tao Xu, Sarah Yoo, Kevin Yu, Qiming Yuan, Wojciech Zaremba, Rowan Zellers, Chong
  Zhang, Marvin Zhang, Shengjia Zhao, Tianhao Zheng, Juntang Zhuang, William Zhuk, and Barret Zoph. 2023.
\newblock \href {https://api.semanticscholar.org/CorpusID:257532815} {Gpt-4 technical report}.

\bibitem[{Akhtar et~al.(2024)Akhtar, Pang, Marzoca, Altun, and Eisenschlos}]{Akhtar2024TANQAO}
Mubashara Akhtar, Chenxi Pang, Andreea Marzoca, Yasemin Altun, and Julian~Martin Eisenschlos. 2024.
\newblock \href {https://api.semanticscholar.org/CorpusID:269757585} {Tanq: An open domain dataset of table answered questions}.

\bibitem[{AlKhamissi et~al.(2022)AlKhamissi, Li, Celikyilmaz, Diab, and Ghazvininejad}]{AlKhamissi2022ARO}
Badr AlKhamissi, Millicent Li, Asli Celikyilmaz, Mona~T. Diab, and Marjan Ghazvininejad. 2022.
\newblock \href {https://api.semanticscholar.org/CorpusID:248157206} {A review on language models as knowledge bases}.
\newblock \emph{ArXiv}, abs/2204.06031.

\bibitem[{Amouyal et~al.(2022)Amouyal, Wolfson, Rubin, Yoran, Herzig, and Berant}]{Amouyal2022QAMPARIA}
Samuel~Joseph Amouyal, Tomer Wolfson, Ohad Rubin, Ori Yoran, Jonathan Herzig, and Jonathan Berant. 2022.
\newblock \href {https://api.semanticscholar.org/CorpusID:265038095} {Qampari: : An open-domain question answering benchmark for questions with many answers from multiple paragraphs}.
\newblock \emph{ArXiv}, abs/2205.12665.

\bibitem[{Berant et~al.(2018)Berant, Deutch, Globerson, Milo, and Wolfson}]{Berant2018ExplainingQO}
Jonathan Berant, Daniel Deutch, Amir Globerson, Tova Milo, and Tomer Wolfson. 2018.
\newblock \href {https://api.semanticscholar.org/CorpusID:52005329} {Explaining queries over web tables to non-experts}.
\newblock \emph{2019 IEEE 35th International Conference on Data Engineering (ICDE)}, pages 1570--1573.

\bibitem[{Bhagavatula et~al.(2015)Bhagavatula, Noraset, and Downey}]{Bhagavatula2015TabELEL}
Chandra Bhagavatula, Thanapon Noraset, and Doug Downey. 2015.
\newblock \href {https://api.semanticscholar.org/CorpusID:14265783} {Tabel: Entity linking in web tables}.
\newblock In \emph{International Workshop on the Semantic Web}.

\bibitem[{Borisov et~al.(2022)Borisov, Se{\ss}ler, Leemann, Pawelczyk, and Kasneci}]{Borisov2022LanguageMA}
Vadim Borisov, Kathrin Se{\ss}ler, Tobias Leemann, Martin Pawelczyk, and Gjergji Kasneci. 2022.
\newblock \href {https://api.semanticscholar.org/CorpusID:252846328} {Language models are realistic tabular data generators}.
\newblock \emph{ArXiv}, abs/2210.06280.

\bibitem[{Brown et~al.(2020)Brown, Mann, Ryder, Subbiah, Kaplan, Dhariwal, Neelakantan, Shyam, Sastry, Askell, Agarwal, Herbert{-}Voss, Krueger, Henighan, Child, Ramesh, Ziegler, Wu, Winter, Hesse, Chen, Sigler, Litwin, Gray, Chess, Clark, Berner, McCandlish, Radford, Sutskever, and Amodei}]{NEURIPS2020_1457c0d6}
Tom~B. Brown, Benjamin Mann, Nick Ryder, Melanie Subbiah, Jared Kaplan, Prafulla Dhariwal, Arvind Neelakantan, Pranav Shyam, Girish Sastry, Amanda Askell, Sandhini Agarwal, Ariel Herbert{-}Voss, Gretchen Krueger, Tom Henighan, Rewon Child, Aditya Ramesh, Daniel~M. Ziegler, Jeffrey Wu, Clemens Winter, Christopher Hesse, Mark Chen, Eric Sigler, Mateusz Litwin, Scott Gray, Benjamin Chess, Jack Clark, Christopher Berner, Sam McCandlish, Alec Radford, Ilya Sutskever, and Dario Amodei. 2020.
\newblock \href {https://proceedings.neurips.cc/paper/2020/hash/1457c0d6bfcb4967418bfb8ac142f64a-Abstract.html} {Language models are few-shot learners}.
\newblock In \emph{Advances in Neural Information Processing Systems 33: Annual Conference on Neural Information Processing Systems 2020, NeurIPS 2020, December 6-12, 2020, virtual}.

\bibitem[{Carlini et~al.(2022)Carlini, Ippolito, Jagielski, Lee, Tram{\`e}r, and Zhang}]{Carlini2022QuantifyingMA}
Nicholas Carlini, Daphne Ippolito, Matthew Jagielski, Katherine Lee, Florian Tram{\`e}r, and Chiyuan Zhang. 2022.
\newblock \href {https://api.semanticscholar.org/CorpusID:246863735} {Quantifying memorization across neural language models}.
\newblock \emph{ArXiv}, abs/2202.07646.

\bibitem[{Chen et~al.(2020{\natexlab{a}})Chen, Chang, Schlinger, Wang, and Cohen}]{Chen2020OpenQA}
Wenhu Chen, Ming-Wei Chang, Eva Schlinger, William~Yang Wang, and William~W. Cohen. 2020{\natexlab{a}}.
\newblock \href {https://api.semanticscholar.org/CorpusID:224803601} {Open question answering over tables and text}.
\newblock \emph{ArXiv}, abs/2010.10439.

\bibitem[{Chen et~al.(2020{\natexlab{b}})Chen, Zha, Chen, Xiong, Wang, and Wang}]{Chen2020HybridQAAD}
Wenhu Chen, Hanwen Zha, Zhiyu Chen, Wenhan Xiong, Hong Wang, and William~Yang Wang. 2020{\natexlab{b}}.
\newblock \href {https://api.semanticscholar.org/CorpusID:215785913} {Hybridqa: A dataset of multi-hop question answering over tabular and textual data}.
\newblock In \emph{Findings}.

\bibitem[{Chen et~al.(2021)Chen, Chen, Smiley, Shah, Borova, Langdon, Moussa, Beane, Huang, Routledge, and Wang}]{Chen2021FinQAAD}
Zhiyu Chen, Wenhu Chen, Charese Smiley, Sameena Shah, Iana Borova, Dylan Langdon, Reema~N Moussa, Matthew~I. Beane, Ting-Hao~'Kenneth' Huang, Bryan~R. Routledge, and William~Yang Wang. 2021.
\newblock \href {https://api.semanticscholar.org/CorpusID:235399966} {Finqa: A dataset of numerical reasoning over financial data}.
\newblock \emph{ArXiv}, abs/2109.00122.

\bibitem[{Chen et~al.(2022)Chen, Li, Smiley, Ma, Shah, and Wang}]{chen-etal-2022-convfinqa}
Zhiyu Chen, Shiyang Li, Charese Smiley, Zhiqiang Ma, Sameena Shah, and William~Yang Wang. 2022.
\newblock \href {https://aclanthology.org/2022.emnlp-main.421} {{C}onv{F}in{QA}: Exploring the chain of numerical reasoning in conversational finance question answering}.
\newblock In \emph{Proceedings of the 2022 Conference on Empirical Methods in Natural Language Processing}, pages 6279--6292, Abu Dhabi, United Arab Emirates. Association for Computational Linguistics.

\bibitem[{Chowdhery et~al.(2022)Chowdhery, Narang, Devlin, Bosma, Mishra, Roberts, Barham, Chung, Sutton, Gehrmann, Schuh, Shi, Tsvyashchenko, Maynez, Rao, Barnes, Tay, Shazeer, Prabhakaran, Reif, Du, Hutchinson, Pope, Bradbury, Austin, Isard, Gur-Ari, Yin, Duke, Levskaya, Ghemawat, Dev, Michalewski, Garc{\'i}a, Misra, Robinson, Fedus, Zhou, Ippolito, Luan, Lim, Zoph, Spiridonov, Sepassi, Dohan, Agrawal, Omernick, Dai, Pillai, Pellat, Lewkowycz, Moreira, Child, Polozov, Lee, Zhou, Wang, Saeta, D{\'i}az, Firat, Catasta, Wei, Meier-Hellstern, Eck, Dean, Petrov, and Fiedel}]{Chowdhery2022PaLMSL}
Aakanksha Chowdhery, Sharan Narang, Jacob Devlin, Maarten Bosma, Gaurav Mishra, Adam Roberts, Paul Barham, Hyung~Won Chung, Charles Sutton, Sebastian Gehrmann, Parker Schuh, Kensen Shi, Sasha Tsvyashchenko, Joshua Maynez, Abhishek Rao, Parker Barnes, Yi~Tay, Noam~M. Shazeer, Vinodkumar Prabhakaran, Emily Reif, Nan Du, Benton~C. Hutchinson, Reiner Pope, James Bradbury, Jacob Austin, Michael Isard, Guy Gur-Ari, Pengcheng Yin, Toju Duke, Anselm Levskaya, Sanjay Ghemawat, Sunipa Dev, Henryk Michalewski, Xavier Garc{\'i}a, Vedant Misra, Kevin Robinson, Liam Fedus, Denny Zhou, Daphne Ippolito, David Luan, Hyeontaek Lim, Barret Zoph, Alexander Spiridonov, Ryan Sepassi, David Dohan, Shivani Agrawal, Mark Omernick, Andrew~M. Dai, Thanumalayan~Sankaranarayana Pillai, Marie Pellat, Aitor Lewkowycz, Erica Moreira, Rewon Child, Oleksandr Polozov, Katherine Lee, Zongwei Zhou, Xuezhi Wang, Brennan Saeta, Mark D{\'i}az, Orhan Firat, Michele Catasta, Jason Wei, Kathleen~S. Meier-Hellstern, Douglas Eck, Jeff Dean, Slav Petrov,
  and Noah Fiedel. 2022.
\newblock Palm: Scaling language modeling with pathways.
\newblock \emph{ArXiv}, abs/2204.02311.

\bibitem[{Codd(1990)}]{codd1990relational}
Edgar~F Codd. 1990.
\newblock \emph{The relational model for database management: version 2}.
\newblock Addison-Wesley Longman Publishing Co., Inc.

\bibitem[{Cohen et~al.(2023)Cohen, Geva, Berant, and Globerson}]{Cohen2023CrawlingTI}
Roi Cohen, Mor Geva, Jonathan Berant, and Amir Globerson. 2023.
\newblock \href {https://api.semanticscholar.org/CorpusID:256389395} {Crawling the internal knowledge-base of language models}.
\newblock In \emph{Findings}.

\bibitem[{Deng et~al.(2021)Deng, Awadallah, Meek, Polozov, Sun, and Richardson}]{Deng2021StructureGroundedPF}
Xiang Deng, Ahmed~Hassan Awadallah, Christopher Meek, Oleksandr Polozov, Huan Sun, and Matthew Richardson. 2021.
\newblock \href {https://doi.org/10.18653/v1/2021.naacl-main.105} {Structure-grounded pretraining for text-to-{SQL}}.
\newblock In \emph{Proceedings of the 2021 Conference of the North American Chapter of the Association for Computational Linguistics: Human Language Technologies}, pages 1337--1350, Online. Association for Computational Linguistics.

\bibitem[{Elazar et~al.(2023)Elazar, Bhagia, Magnusson, Ravichander, Schwenk, Suhr, Walsh, Groeneveld, Soldaini, Singh, Hajishirzi, Smith, and Dodge}]{Elazar2023WhatsIM}
Yanai Elazar, Akshita Bhagia, Ian~H. Magnusson, Abhilasha Ravichander, Dustin Schwenk, Alane Suhr, Pete Walsh, Dirk Groeneveld, Luca Soldaini, Sameer Singh, Hanna Hajishirzi, Noah~A. Smith, and Jesse Dodge. 2023.
\newblock \href {https://api.semanticscholar.org/CorpusID:264803575} {What's in my big data?}
\newblock \emph{ArXiv}, abs/2310.20707.

\bibitem[{Fang et~al.(2024)Fang, Xu, Tan, Zhang, Hu, Qi, Nickleach, Socolinsky, Sengamedu, and Faloutsos}]{Fang2024LargeLM}
Xi~Fang, Weijie Xu, Fiona~Anting Tan, Jiani Zhang, Ziqing Hu, Yanjun Qi, Scott Nickleach, Diego Socolinsky, Srinivasan Sengamedu, and Christos Faloutsos. 2024.
\newblock \href {https://api.semanticscholar.org/CorpusID:268041519} {Large language models(llms) on tabular data: Prediction, generation, and understanding - a survey}.
\newblock \emph{ArXiv}, abs/2402.17944.

\bibitem[{Johnson et~al.(2016)Johnson, Pollard, Shen, wei H.~Lehman, Feng, Ghassemi, Moody, Szolovits, Celi, and Mark}]{Johnson2016MIMICIIIAF}
Alistair E.~W. Johnson, Tom~J. Pollard, Lu~Shen, Li~wei H.~Lehman, Mengling Feng, Mohammad~Mahdi Ghassemi, Benjamin Moody, Peter Szolovits, Leo~Anthony Celi, and Roger~G. Mark. 2016.
\newblock \href {https://api.semanticscholar.org/CorpusID:33285731} {Mimic-iii, a freely accessible critical care database}.
\newblock \emph{Scientific Data}, 3.

\bibitem[{Kadavath et~al.(2022)Kadavath, Conerly, Askell, Henighan, Drain, Perez, Schiefer, Dodds, DasSarma, Tran-Johnson, Johnston, El-Showk, Jones, Elhage, Hume, Chen, Bai, Bowman, Fort, Ganguli, Hernandez, Jacobson, Kernion, Kravec, Lovitt, Ndousse, Olsson, Ringer, Amodei, Brown, Clark, Joseph, Mann, McCandlish, Olah, and Kaplan}]{Kadavath2022LanguageM}
Saurav Kadavath, Tom Conerly, Amanda Askell, T.~J. Henighan, Dawn Drain, Ethan Perez, Nicholas Schiefer, Zachary Dodds, Nova DasSarma, Eli Tran-Johnson, Scott Johnston, Sheer El-Showk, Andy Jones, Nelson Elhage, Tristan Hume, Anna Chen, Yuntao Bai, Sam Bowman, Stanislav Fort, Deep Ganguli, Danny Hernandez, Josh Jacobson, John Kernion, Shauna Kravec, Liane Lovitt, Kamal Ndousse, Catherine Olsson, Sam Ringer, Dario Amodei, Tom~B. Brown, Jack Clark, Nicholas Joseph, Benjamin Mann, Sam McCandlish, Christopher Olah, and Jared Kaplan. 2022.
\newblock Language models (mostly) know what they know.
\newblock \emph{ArXiv}, abs/2207.05221.

\bibitem[{Khot et~al.(2022)Khot, Trivedi, Finlayson, Fu, Richardson, Clark, and Sabharwal}]{khot2022decomposed}
Tushar Khot, Harsh Trivedi, Matthew Finlayson, Yao Fu, Kyle Richardson, Peter Clark, and Ashish Sabharwal. 2022.
\newblock \href {https://arxiv.org/abs/2210.02406} {Decomposed prompting: A modular approach for solving complex tasks}.
\newblock \emph{ArXiv preprint}, abs/2210.02406.

\bibitem[{Khot et~al.(2023)Khot, Trivedi, Finlayson, Fu, Richardson, Clark, and Sabharwal}]{khot2023decomposed}
Tushar Khot, Harsh Trivedi, Matthew Finlayson, Yao Fu, Kyle Richardson, Peter Clark, and Ashish Sabharwal. 2023.
\newblock \href {http://arxiv.org/abs/2210.02406} {Decomposed prompting: A modular approach for solving complex tasks}.

\bibitem[{Kwiatkowski et~al.(2019)Kwiatkowski, Palomaki, Redfield, Collins, Parikh, Alberti, Epstein, Polosukhin, Devlin, Lee, Toutanova, Jones, Kelcey, Chang, Dai, Uszkoreit, Le, and Petrov}]{kwiatkowski-etal-2019-natural}
Tom Kwiatkowski, Jennimaria Palomaki, Olivia Redfield, Michael Collins, Ankur Parikh, Chris Alberti, Danielle Epstein, Illia Polosukhin, Jacob Devlin, Kenton Lee, Kristina Toutanova, Llion Jones, Matthew Kelcey, Ming-Wei Chang, Andrew~M. Dai, Jakob Uszkoreit, Quoc Le, and Slav Petrov. 2019.
\newblock \href {https://doi.org/10.1162/tacl_a_00276} {Natural questions: A benchmark for question answering research}.
\newblock \emph{Transactions of the Association for Computational Linguistics}, 7:452--466.

\bibitem[{Lewis et~al.(2020)Lewis, Perez, Piktus, Petroni, Karpukhin, Goyal, Kuttler, Lewis, tau Yih, Rockt{\"a}schel, Riedel, and Kiela}]{Lewis2020RetrievalAugmentedGF}
Patrick Lewis, Ethan Perez, Aleksandara Piktus, Fabio Petroni, Vladimir Karpukhin, Naman Goyal, Heinrich Kuttler, Mike Lewis, Wen tau Yih, Tim Rockt{\"a}schel, Sebastian Riedel, and Douwe Kiela. 2020.
\newblock \href {https://api.semanticscholar.org/CorpusID:218869575} {Retrieval-augmented generation for knowledge-intensive nlp tasks}.
\newblock \emph{ArXiv}, abs/2005.11401.

\bibitem[{Li et~al.(2023)Li, He, Yashar, Cui, Ge, Zhang, Fainman, Zhang, and Chaudhuri}]{Li2023TableGPTTG}
Peng Li, Yeye He, Dror Yashar, Weiwei Cui, Song Ge, Haidong Zhang, Danielle~Rifinski Fainman, Dongmei Zhang, and Surajit Chaudhuri. 2023.
\newblock \href {https://api.semanticscholar.org/CorpusID:264127877} {Table-gpt: Table-tuned gpt for diverse table tasks}.
\newblock \emph{ArXiv}, abs/2310.09263.

\bibitem[{Malaviya et~al.(2023)Malaviya, Shaw, Chang, Lee, and Toutanova}]{Malaviya2023QUESTAR}
Chaitanya Malaviya, Peter Shaw, Ming-Wei Chang, Kenton Lee, and Kristina Toutanova. 2023.
\newblock \href {https://api.semanticscholar.org/CorpusID:258822815} {Quest: A retrieval dataset of entity-seeking queries with implicit set operations}.
\newblock \emph{ArXiv}, abs/2305.11694.

\bibitem[{Mallen et~al.(2022)Mallen, Asai, Zhong, Das, Hajishirzi, and Khashabi}]{Mallen2022WhenNT}
Alex~Troy Mallen, Akari Asai, Victor Zhong, Rajarshi Das, Hannaneh Hajishirzi, and Daniel Khashabi. 2022.
\newblock \href {https://api.semanticscholar.org/CorpusID:254877603} {When not to trust language models: Investigating effectiveness of parametric and non-parametric memories}.
\newblock In \emph{Annual Meeting of the Association for Computational Linguistics}.

\bibitem[{Nan et~al.(2022)Nan, Hsieh, Mao, Lin, Verma, Zhang, Kry{\'s}ci{\'n}ski, Schoelkopf, Kong, Tang, Mutuma, Rosand, Trindade, Bandaru, Cunningham, Xiong, Radev, and Radev}]{nan-etal-2022-fetaqa}
Linyong Nan, Chiachun Hsieh, Ziming Mao, Xi~Victoria Lin, Neha Verma, Rui Zhang, Wojciech Kry{\'s}ci{\'n}ski, Hailey Schoelkopf, Riley Kong, Xiangru Tang, Mutethia Mutuma, Ben Rosand, Isabel Trindade, Renusree Bandaru, Jacob Cunningham, Caiming Xiong, Dragomir Radev, and Dragomir Radev. 2022.
\newblock \href {https://doi.org/10.1162/tacl_a_00446} {{F}e{T}a{QA}: Free-form table question answering}.
\newblock \emph{Transactions of the Association for Computational Linguistics}, 10:35--49.

\bibitem[{Pal et~al.(2023)Pal, Yates, Kanoulas, and de~Rijke}]{Pal2023MultiTabQAGT}
Vaishali Pal, Andrew Yates, E.~Kanoulas, and M.~de~Rijke. 2023.
\newblock \href {https://api.semanticscholar.org/CorpusID:258833465} {Multitabqa: Generating tabular answers for multi-table question answering}.
\newblock In \emph{Annual Meeting of the Association for Computational Linguistics}.

\bibitem[{Parikh et~al.(2020)Parikh, Wang, Gehrmann, Faruqui, Dhingra, Yang, and Das}]{Parikh2020ToTToAC}
Ankur Parikh, Xuezhi Wang, Sebastian Gehrmann, Manaal Faruqui, Bhuwan Dhingra, Diyi Yang, and Dipanjan Das. 2020.
\newblock \href {https://doi.org/10.18653/v1/2020.emnlp-main.89} {{ToTTo}: A controlled table-to-text generation dataset}.
\newblock In \emph{Proceedings of the 2020 Conference on Empirical Methods in Natural Language Processing (EMNLP)}, pages 1173--1186, Online. Association for Computational Linguistics.

\bibitem[{Pasupat and Liang(2015)}]{Pasupat2015CompositionalSP}
Panupong Pasupat and Percy Liang. 2015.
\newblock \href {https://api.semanticscholar.org/CorpusID:9027681} {Compositional semantic parsing on semi-structured tables}.
\newblock In \emph{Annual Meeting of the Association for Computational Linguistics}.

\bibitem[{Petroni et~al.(2019)Petroni, Rockt{\"a}schel, Riedel, Lewis, Bakhtin, Wu, and Miller}]{petroni2019language}
Fabio Petroni, Tim Rockt{\"a}schel, Sebastian Riedel, Patrick Lewis, Anton Bakhtin, Yuxiang Wu, and Alexander Miller. 2019.
\newblock \href {https://doi.org/10.18653/v1/D19-1250} {Language models as knowledge bases?}
\newblock In \emph{Proceedings of the 2019 Conference on Empirical Methods in Natural Language Processing and the 9th International Joint Conference on Natural Language Processing (EMNLP-IJCNLP)}, pages 2463--2473, Hong Kong, China. Association for Computational Linguistics.

\bibitem[{Seedat et~al.(2023)Seedat, Huynh, van Breugel, and van~der Schaar}]{Seedat2023CuratedLS}
Nabeel Seedat, Nicolas Huynh, Boris van Breugel, and Mihaela van~der Schaar. 2023.
\newblock \href {https://api.semanticscholar.org/CorpusID:266362265} {Curated llm: Synergy of llms and data curation for tabular augmentation in ultra low-data regimes}.
\newblock \emph{ArXiv}, abs/2312.12112.

\bibitem[{Shaham et~al.(2022)Shaham, Segal, Ivgi, Efrat, Yoran, Haviv, Gupta, Xiong, Geva, Berant, and Levy}]{shaham-etal-2022-scrolls}
Uri Shaham, Elad Segal, Maor Ivgi, Avia Efrat, Ori Yoran, Adi Haviv, Ankit Gupta, Wenhan Xiong, Mor Geva, Jonathan Berant, and Omer Levy. 2022.
\newblock \href {https://aclanthology.org/2022.emnlp-main.823} {{SCROLLS}: Standardized {C}ompa{R}ison over long language sequences}.
\newblock In \emph{Proceedings of the 2022 Conference on Empirical Methods in Natural Language Processing}, pages 12007--12021, Abu Dhabi, United Arab Emirates. Association for Computational Linguistics.

\bibitem[{Shao et~al.(2024)Shao, Jiang, Kanell, Xu, Khattab, and Lam}]{Shao2024AssistingIW}
Yijia Shao, Yucheng Jiang, Theodore~A. Kanell, Peter Xu, Omar Khattab, and Monica~S. Lam. 2024.
\newblock \href {https://api.semanticscholar.org/CorpusID:267782917} {Assisting in writing wikipedia-like articles from scratch with large language models}.
\newblock \emph{ArXiv}, abs/2402.14207.

\bibitem[{Shen et~al.(2021)Shen, Shen, Luo, Yang, Hu, Zhang, Tai, and Wang}]{Shen2021TowardsNL}
Leixian Shen, Enya Shen, Yuyu Luo, Xiaocong Yang, Xuming Hu, Xiongshuai Zhang, Zhiwei Tai, and Jianmin Wang. 2021.
\newblock \href {https://api.semanticscholar.org/CorpusID:237439677} {Towards natural language interfaces for data visualization: A survey}.
\newblock \emph{IEEE Transactions on Visualization and Computer Graphics}, 29:3121--3144.

\bibitem[{Singha et~al.(2023)Singha, Cambronero, Gulwani, Le, and Parnin}]{singha2023tabular}
Ananya Singha, José Cambronero, Sumit Gulwani, Vu~Le, and Chris Parnin. 2023.
\newblock \href {http://arxiv.org/abs/2310.10358} {Tabular representation, noisy operators, and impacts on table structure understanding tasks in llms}.

\bibitem[{Sui et~al.(2023)Sui, Zhou, Zhou, Han, and Zhang}]{Sui2023TableML}
Yuan Sui, Mengyu Zhou, Mingjie Zhou, Shi Han, and Dongmei Zhang. 2023.
\newblock \href {https://api.semanticscholar.org/CorpusID:258833533} {Table meets llm: Can large language models understand structured table data? a benchmark and empirical study}.
\newblock In \emph{Web Search and Data Mining}.

\bibitem[{Touvron et~al.(2023{\natexlab{a}})Touvron, Lavril, Izacard, Martinet, Lachaux, Lacroix, Rozi{\`e}re, Goyal, Hambro, Azhar, Rodriguez, Joulin, Grave, and Lample}]{Touvron2023LLaMAOA}
Hugo Touvron, Thibaut Lavril, Gautier Izacard, Xavier Martinet, Marie-Anne Lachaux, Timoth{\'e}e Lacroix, Baptiste Rozi{\`e}re, Naman Goyal, Eric Hambro, Faisal Azhar, Aur'elien Rodriguez, Armand Joulin, Edouard Grave, and Guillaume Lample. 2023{\natexlab{a}}.
\newblock Llama: Open and efficient foundation language models.
\newblock \emph{ArXiv}, abs/2302.13971.

\bibitem[{Touvron et~al.(2023{\natexlab{b}})Touvron, Martin, Stone, Albert, Almahairi, Babaei, Bashlykov, Batra, Bhargava, Bhosale, Bikel, Blecher, Ferrer, Chen, Cucurull, Esiobu, Fernandes, Fu, Fu, Fuller, Gao, Goswami, Goyal, Hartshorn, Hosseini, Hou, Inan, Kardas, Kerkez, Khabsa, Kloumann, Korenev, Koura, Lachaux, Lavril, Lee, Liskovich, Lu, Mao, Martinet, Mihaylov, Mishra, Molybog, Nie, Poulton, Reizenstein, Rungta, Saladi, Schelten, Silva, Smith, Subramanian, Tan, Tang, Taylor, Williams, Kuan, Xu, Yan, Zarov, Zhang, Fan, Kambadur, Narang, Rodriguez, Stojnic, Edunov, and Scialom}]{Touvron2023Llama2O}
Hugo Touvron, Louis Martin, Kevin~R. Stone, Peter Albert, Amjad Almahairi, Yasmine Babaei, Nikolay Bashlykov, Soumya Batra, Prajjwal Bhargava, Shruti Bhosale, Daniel~M. Bikel, Lukas Blecher, Cristian~Cant{\'o}n Ferrer, Moya Chen, Guillem Cucurull, David Esiobu, Jude Fernandes, Jeremy Fu, Wenyin Fu, Brian Fuller, Cynthia Gao, Vedanuj Goswami, Naman Goyal, Anthony~S. Hartshorn, Saghar Hosseini, Rui Hou, Hakan Inan, Marcin Kardas, Viktor Kerkez, Madian Khabsa, Isabel~M. Kloumann, A.~V. Korenev, Punit~Singh Koura, Marie-Anne Lachaux, Thibaut Lavril, Jenya Lee, Diana Liskovich, Yinghai Lu, Yuning Mao, Xavier Martinet, Todor Mihaylov, Pushkar Mishra, Igor Molybog, Yixin Nie, Andrew Poulton, Jeremy Reizenstein, Rashi Rungta, Kalyan Saladi, Alan Schelten, Ruan Silva, Eric~Michael Smith, R.~Subramanian, Xia Tan, Binh Tang, Ross Taylor, Adina Williams, Jian~Xiang Kuan, Puxin Xu, Zhengxu Yan, Iliyan Zarov, Yuchen Zhang, Angela Fan, Melanie Kambadur, Sharan Narang, Aurelien Rodriguez, Robert Stojnic, Sergey Edunov, and
  Thomas Scialom. 2023{\natexlab{b}}.
\newblock \href {https://api.semanticscholar.org/CorpusID:259950998} {Llama 2: Open foundation and fine-tuned chat models}.
\newblock \emph{ArXiv}, abs/2307.09288.

\bibitem[{Wolfson et~al.(2022)Wolfson, Deutch, and Berant}]{wolfson-etal-2022-weakly}
Tomer Wolfson, Daniel Deutch, and Jonathan Berant. 2022.
\newblock \href {https://doi.org/10.18653/v1/2022.findings-naacl.193} {Weakly supervised text-to-{SQL} parsing through question decomposition}.
\newblock In \emph{Findings of the Association for Computational Linguistics: NAACL 2022}, pages 2528--2542, Seattle, United States. Association for Computational Linguistics.

\bibitem[{Yoran et~al.(2023)Yoran, Wolfson, Ram, and Berant}]{yoran2023making}
Ori Yoran, Tomer Wolfson, Ori Ram, and Jonathan Berant. 2023.
\newblock \href {http://arxiv.org/abs/2310.01558} {Making retrieval-augmented language models robust to irrelevant context}.

\bibitem[{Yu et~al.(2018)Yu, Zhang, Yang, Yasunaga, Wang, Li, Ma, Li, Yao, Roman, Zhang, and Radev}]{Yu2018SpiderAL}
Tao Yu, Rui Zhang, Kai Yang, Michihiro Yasunaga, Dongxu Wang, Zifan Li, James Ma, Irene Li, Qingning Yao, Shanelle Roman, Zilin Zhang, and Dragomir Radev. 2018.
\newblock \href {https://doi.org/10.18653/v1/D18-1425} {{S}pider: A large-scale human-labeled dataset for complex and cross-domain semantic parsing and text-to-{SQL} task}.
\newblock In \emph{Proceedings of the 2018 Conference on Empirical Methods in Natural Language Processing}, pages 3911--3921, Brussels, Belgium. Association for Computational Linguistics.

\bibitem[{Zhang and Choi(2021)}]{zhang-choi-2021-situatedqa}
Michael Zhang and Eunsol Choi. 2021.
\newblock \href {https://doi.org/10.18653/v1/2021.emnlp-main.586} {{S}ituated{QA}: Incorporating extra-linguistic contexts into {QA}}.
\newblock In \emph{Proceedings of the 2021 Conference on Empirical Methods in Natural Language Processing}, pages 7371--7387, Online and Punta Cana, Dominican Republic. Association for Computational Linguistics.

\bibitem[{Zhang et~al.(2023)Zhang, Wang, Yan, Li, and Liu}]{Zhang2023GenerativeTP}
T.~Zhang, Shaowen Wang, Shuicheng Yan, Jian Li, and Qian Liu. 2023.
\newblock \href {https://api.semanticscholar.org/CorpusID:258740999} {Generative table pre-training empowers models for tabular prediction}.
\newblock \emph{ArXiv}, abs/2305.09696.

\end{thebibliography}
\bibliographystyle{acl_natbib}

\newpage
\appendix 

\section{Table Generation Prompts}
\label{appendix:prompts}

In this section we provide the prompt templates used in each of our table generation methods. Figs.~\ref{fig:full_table}-\ref{fig:cells_generation} present our prompt templates used for: full table generation method, keys generation, row-by-row, and cell-by-cell method.

\begin{figure*}[t]\setlength{\belowcaptionskip}{-8pt}
\footnotesize
  \begin{tabular}{p{0.97\linewidth}}
    \toprule \textbf{Full-table generation template:}\\
You are a retriever of facts. List all \texttt{\{table description\}}.
    The response will be formatted as JSON shown below.
    Each element of the response will contain \texttt{\{num columns\}} fields: \{\texttt{column1}, \texttt{column2}, ...\}\\
    Do not output any additional text that is not in JSON format.
    
    RESPONSE FORMAT:
    [\{
        \texttt{column1: value1, column2: value2, ...}
    \}]
\\

\midrule
{ \textbf{Full-table generation (populated example):}}\\
You are a retriever of facts. List all \hl{achievements of Susen Tiedtke from 1987 to 2000}.
    The response will be formatted as JSON shown below.
    Each element of the response will contain \hl{4} fields: \hl{['year', 'competition', 'venue', 'position']}.
    Do not output any additional text that is not in JSON format.
    
    RESPONSE FORMAT:
    [\{
        \hl{``year'': \_year, ``competition'': \_competition, ``venue'': \_venue, ``position'': \_position}
    \}]
\\
    
\bottomrule
\end{tabular}
\caption{Full-table generation prompt.}
\label{fig:full_table}
\end{figure*}

\begin{figure*}[t]\setlength{\belowcaptionskip}{-8pt}
\footnotesize
  \begin{tabular}{p{0.97\linewidth}}
    \toprule \textbf{Keys  generation template:}\\
You are a retriever of facts.
We want to create a table with the detailed information about \texttt{\{table description\}}.
The key columns in the table are \texttt{\texttt{\{key1, (key2, ...)\}}}.
List all \texttt{\texttt{\{key1, (key2, ...)\}}} entities for the table.
    The response will be formatted as JSON list shown below.    
    
    RESPONSE FORMAT:
    [\{
        \texttt{key: value1, key2: value2, ...}
    \}]
\\

\midrule
{ \textbf{Keys generation (populated example):}}\\
You are a retriever of facts.
We want to create a table with the detailed information about \hl{achievements of Susen Tiedtke from 1987 to 2000}.
The key columns in the table are \hl{competition, year}.
List all \hl{competition, year} entities for the table.
    The response will be formatted as JSON list shown below.
        
    RESPONSE FORMAT:
    [\{
        \hl{``competition'': \_competition, ``year'': \_year}
    \}]
\\
    
\bottomrule
\end{tabular}
\caption{Key columns generation prompt.}
\label{fig:keys_generation}
\end{figure*}

\begin{figure*}[t]\setlength{\belowcaptionskip}{-8pt}
\footnotesize
  \begin{tabular}{p{0.97\linewidth}}
    \toprule \textbf{Row  generation template:}\\
You are a retriever of facts.
We want to create a table with the detailed information about \texttt{\{table description\}}.
Columns in the table are \texttt{\{columns\}}.
The key columns in the table are \texttt{\{key1, (key2, ...)\}}.
Retrieve a single row whose key is \texttt{(\{key = value\})}.  
The response will be formatted as JSON dictionary shown below.
Pay special attention to wrap all values in double quotes!
    
    RESPONSE FORMAT:
    [\{
        \texttt{column1: value1, column2: value2, ...}
    \}]
\\

\midrule
{ \textbf{Row generation (populated example):}}\\
You are a retriever of facts.
We want to create a table with the detailed information about \hl{achievements of Susen Tiedtke from 1987 to 2000}.
Columns in the table are \hl{year, competition, venue, position}.
The key columns in the table are \hl{competition, year}.
Retrieve a single row whose key is (\hl{year = 1987, competition = World Championships}).
The response will be formatted as JSON dictionary shown below.
Pay special attention to wrap all values in double quotes!
        
    RESPONSE FORMAT:
    \{
        \hl{``year'': 1987, ``competition'': World Championships, ``venue'': \_venue, ``position'': \_position}
    \}
\\
    
\bottomrule
\end{tabular}
\caption{Row-by-row (row generation) prompt.}
\label{fig:rows_generation}
\end{figure*}

\begin{figure*}[t]\setlength{\belowcaptionskip}{-8pt}
\footnotesize
  \begin{tabular}{p{0.97\linewidth}}
    \toprule \textbf{Cell  generation template:}\\
You are a retriever of facts.
We want to create a table with the detailed information about \texttt{\{table description\}}.
Columns in the table are \{\texttt{column1}, \texttt{column2}, ...\}.
The key columns in the table are \texttt{\{key1, (key2, ...)\}}.
For the table row whose key is is \texttt{(\{key = value\})} what is the value of attribute \texttt{\{column\}}.
The response will be formatted as JSON dictionary shown below.
Pay special attention to wrap all values in double quotes!
    
    RESPONSE FORMAT:
    \{
        \texttt{column: value}
    \}
\\

\midrule
{ \textbf{Cell generation (populated example):}}\\
You are a retriever of facts.
We want to create a table with the detailed information about \hl{achievements of Susen Tiedtke from 1987 to 2000}.
Columns in the table are \hl{year, competition, venue, position}.
The key columns in the table are \hl{competition, year}.
For the table row whose key is (\hl{year = 1987, competition = World Championships}) what is the value of attribute \hl{venue}.
The response will be formatted as JSON dictionary shown below.
Pay special attention to wrap all values in double quotes!
        
    RESPONSE FORMAT:
    \{
        \hl{``venue'': \_venue}
    \}
\\
    
\bottomrule
\end{tabular}
\caption{Cell-by-cell (cell generation) prompt.}
\label{fig:cells_generation}
\end{figure*}

\begin{figure*}[t]
    \centering
    \begin{subfigure}{0.3\textwidth}
        \centering
        \includegraphics[width=\linewidth]{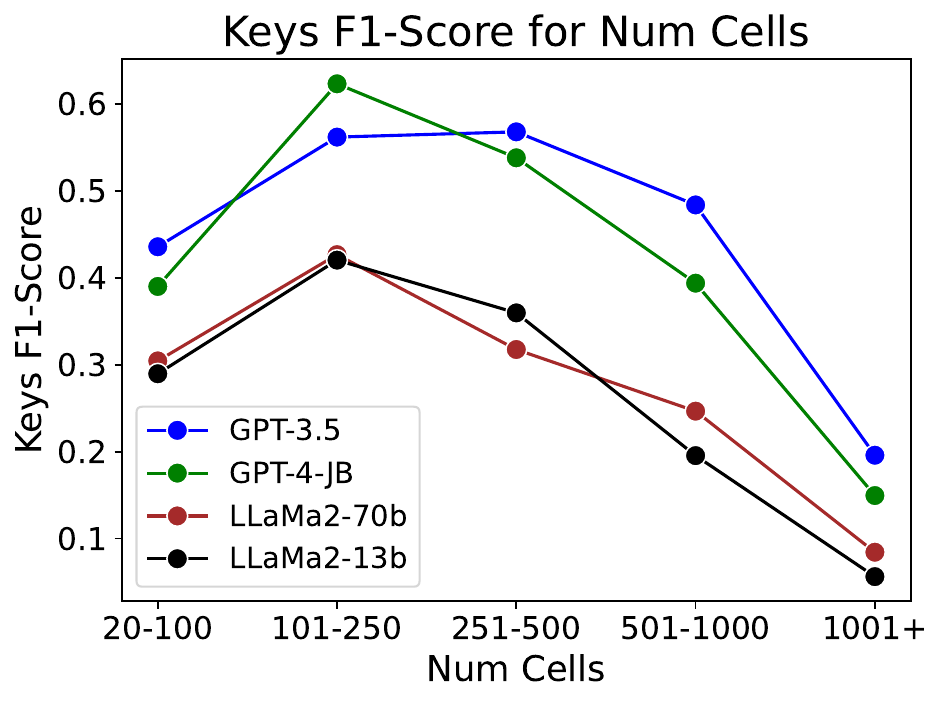}
        \caption{Table Cells Num. - Keys F1}
    \end{subfigure}
    \hfill
    \begin{subfigure}{0.3\textwidth}
        \centering
        \includegraphics[width=\linewidth]{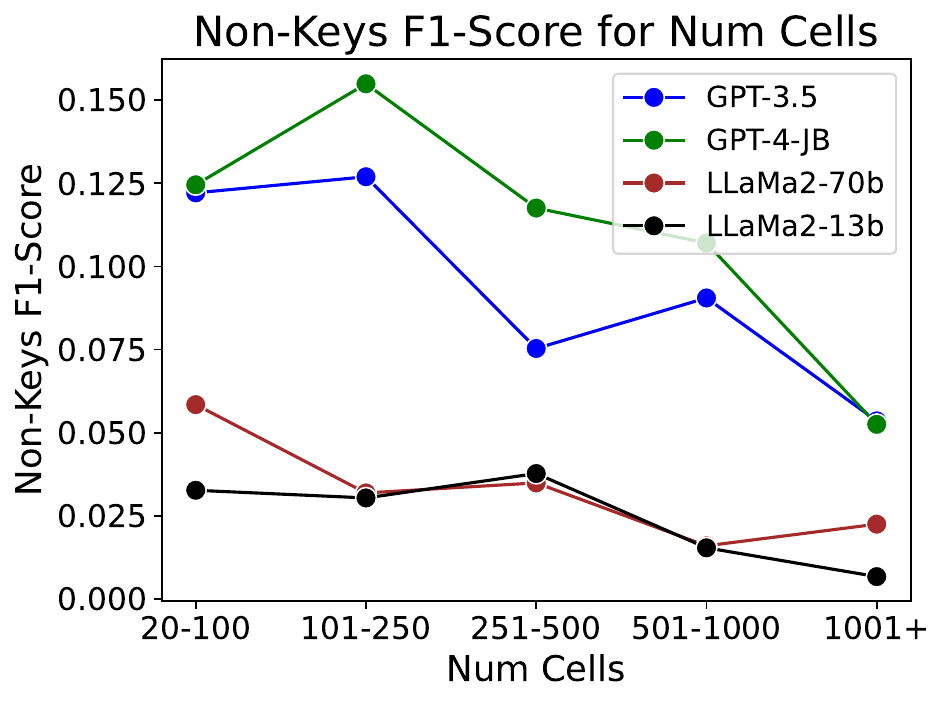}
        \caption{Table Cells Num. - Non-Keys F1}
    \end{subfigure}
    \hfill
    \begin{subfigure}{0.3\textwidth}
        \centering
        \includegraphics[width=\linewidth]{figures/3x3/num_cells_f1_score.pdf}
        \caption{Table Cells Num. - Table F1}
    \end{subfigure}

    \begin{subfigure}{0.3\textwidth}
        \centering
        \includegraphics[width=\linewidth]{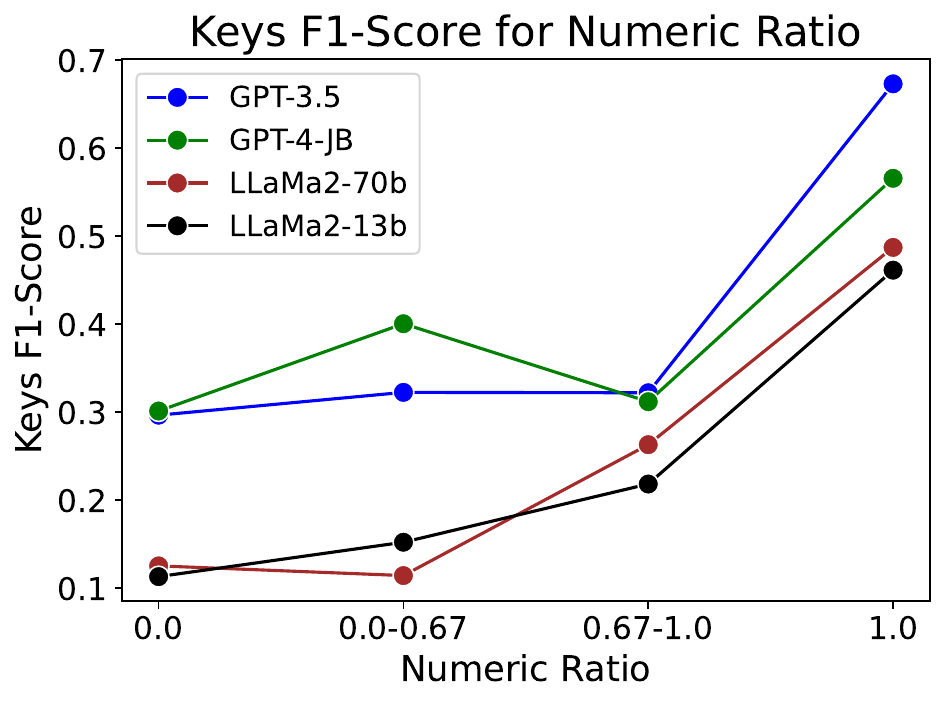}
        \caption{Numeric Cols. \%z - Keys F1}
    \end{subfigure}
    \hfill
    \begin{subfigure}{0.3\textwidth}
        \centering
        \includegraphics[width=\linewidth]{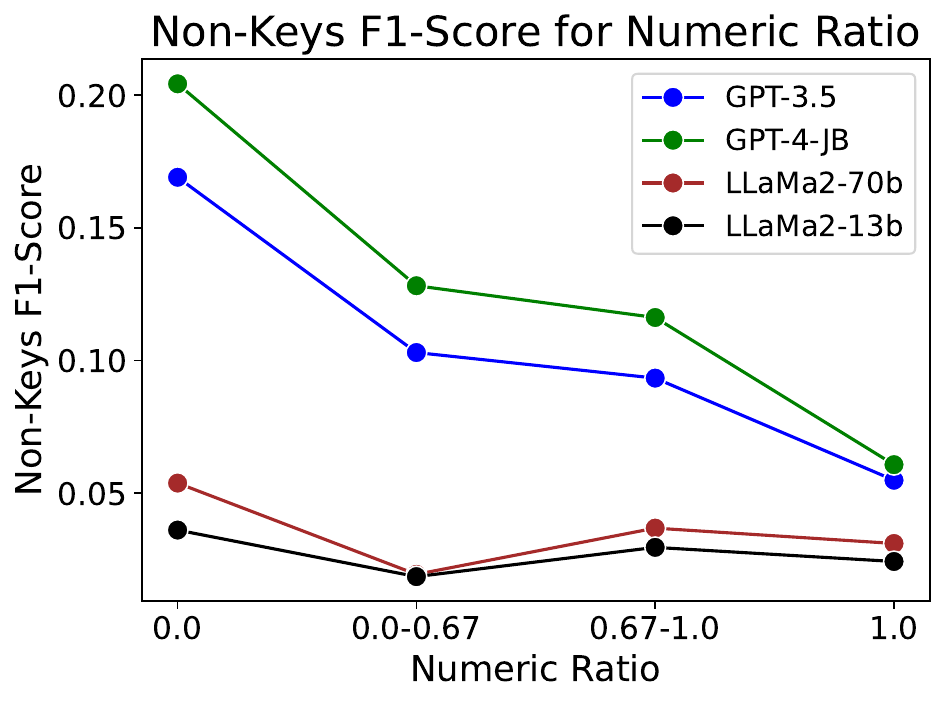}
        \caption{Numeric Cols. \% - Non-Keys F1}
    \end{subfigure}
    \hfill
    \begin{subfigure}{0.3\textwidth}
        \centering
        \includegraphics[width=\linewidth]{figures/3x3/numeric_ratio_f1_score.pdf}
        \caption{Numeric Cols. \% - Table F1}
    \end{subfigure}

    \begin{subfigure}{0.3\textwidth}
        \centering
        \includegraphics[width=\linewidth]{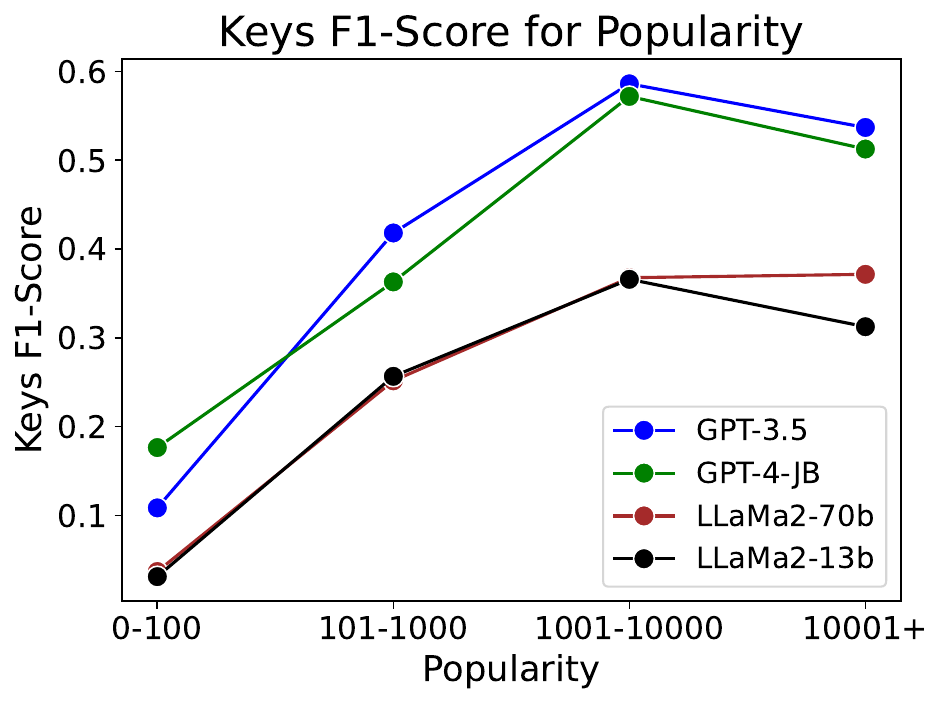}
        \caption{Table Popularity - Keys F1}
    \end{subfigure}
    \hfill
    \begin{subfigure}{0.3\textwidth}
        \centering
        \includegraphics[width=\linewidth]{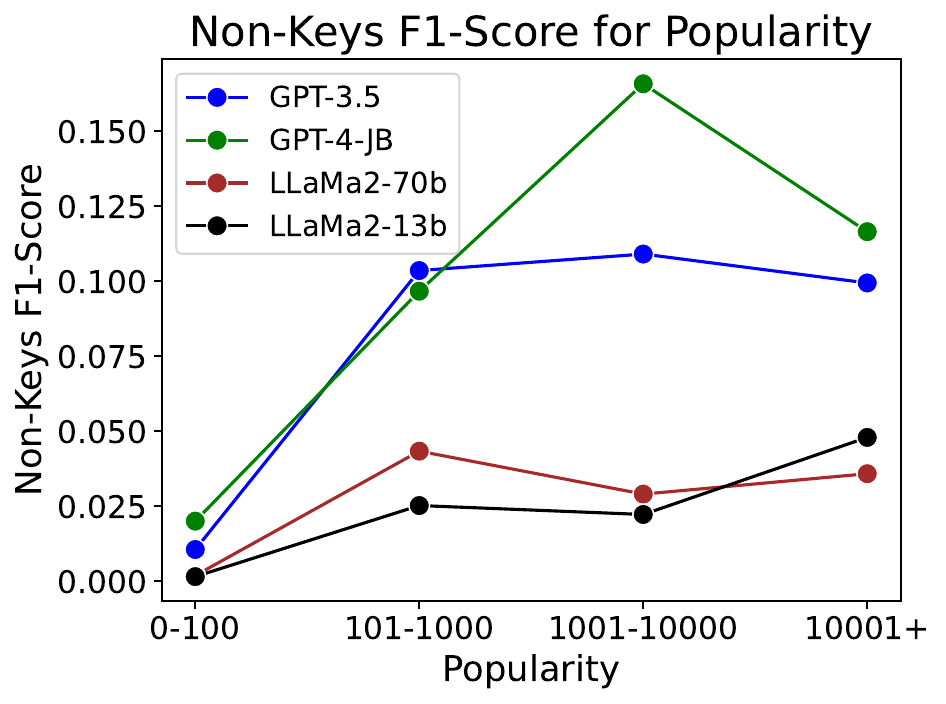}
        \caption{Table Popularity - Non-Keys F1}
    \end{subfigure}
    \hfill
    \begin{subfigure}{0.3\textwidth}
        \centering
        \includegraphics[width=\linewidth]{figures/3x3/popularity_f1_score.pdf}
        \caption{Table Popularity - Table F1}
    \end{subfigure}

    \caption{The effect of table size, the ratio of numeric columns, and table popularity on the generation performance of the full-table method, with four different LLMs. Additional breakdown of generation performance based on cells in key columns versus non-key columns.}
    \label{fig:comparison_full}
\end{figure*}

\section{Evaluation Method Details}
\label{app:fuzzy}

\subsection{Cell Value Matching} 
We next describe our evaluation method in more detail, given an output table $\hat{T}(\hat{R}, C)$ and ground-truth table $T(R, C)$.

As described in \S\ref{ssec:evaluation_method}, we use \textit{exact} value comparison of cell textual content and allow a $\pm 0.1\%$ error for numeric values.
Before comparing textual cells, we first convert them to lower case, and remove non alphanumeric symbols and spaces.

As for date values, we first parse and convert cells with date values to a Python Date object, and then compare the canonical dates. This is to avoid cases where cells are deemed as a non-match due to differences in the date format. For example, in our evaluation process, two date values representing the same date, such as "2014-05-16" and "16th, May, 2014", will be considered the same.  

We further treat ``none'', ``n/a'' ,``nan'' and empty cells as identical in terms of value matching.

\vspace{2mm}
\subsection{Precision and Recall Computation for Tables}
For a given output table $\hat{T}(\hat{R}, C)$ and ground-truth table $T(R, C)$, we first align the rows $\hat{R}$ to their corresponding rows in $R$ by matching their respective keys, namely $\hat{r} \leadsto r \iff \hat{r}[C_k] = r[C_k]$. For rows with composite keys, all key values must be identical, i.e., $\forall c_k \in C_k \hat{r}[c_k] = r[c_k]$.

Recall that a \textit{correct} cell in $T(\hat{R}, C)$ is a cell $\hat{r}[c]$ such that $\hat{r}[c] = r[c] \wedge \hat{r} \leadsto r$. Namely, row $\hat{r}$ is aligned with a row $r$ in the ground-truth table, and their corresponding cell values in column $c$ are identical.

We next provide the precision and recall formulas we used for keys, non-keys, and tables. 

For keys, we compare $\hat{R}[C_k]$ and $R[C_k]$ as follows. Let the number of matching keys $\phi = |\{r \in \hat{R}, \forall c_k \in C_k~\hat{r}[c_k]=r[c_k] \} | $ Then \textit{keys precision} is calculated by $\frac{\phi}{|\hat{R}|}$ and \textit{keys recall} is given by $\frac{\phi}{|{R}|}$. 

For non-keys, we compare $\hat{R}[C \setminus C_k]$ and $R[C \setminus C_k]$. 
After aligning $\hat{R}$ and $R$, we compute the number of \textit{correct} keys, 
denoted by $\psi = | \{(r,c), r\in \hat{R} \wedge c\in C\setminus C_k \wedge r\leadsto \hat{r} \wedge \hat{r}[c]=r[c]    \} |$. 
Then the \textit{non-keys precision} is calculated by $\frac{\psi}{|\hat{R}[C\setminus C_k]|}$ and \textit{non-keys recall} is calculated by $\frac{\psi}{|R[C\setminus C_k]|}$. 

Last, for the table precision and recall, we perform a similar evaluation, now defining the number of correct cells, denoted by $\tau$, as all correct cells in the table. Namely,  $\tau = | \{(r,c), r\in \hat{R} \wedge c\in C \wedge r\leadsto \hat{r} \wedge \hat{r}[c]=r[c]    \} |$, then the \textit{table precision} is simply calculated by $\frac{\tau}{|\hat{R}[C]|}$ and \textit{table recall} is calculated by $\frac{\tau}{|R[C]|}$.

\section{Table Properties Effect on Performance}
In \S\ref{ssec:properties_effects}  we examine how the table properties such as the size, amount of numeric data, and table popularity affect the generation performance. In Fig.~\ref{fig:comparison_full} we present the effect of these three properties on both the keys F1, non-keys F1, and full table F1. 
We can see, for instance, that the table size negatively affects both the keys F1 and the non-keys F1 scores (see Fig.~\ref{fig:comparison_full} (a) and Fig.~\ref{fig:comparison_full} (b)), and the ratio of numeric columns has a negative effect, as expected, only the non-keys F1 (see Fig.~\ref{fig:comparison_full} (e)). 
The table popularity also have a strong effect on both the keys F1 and the non-keys F1 (Fig.~\ref{fig:comparison_full} (g) and Fig.~\ref{fig:comparison_full} (h)).

\label{app:comparison}

\end{document}